\documentclass[11pt]{article}

\usepackage[preprint]{acl}

\usepackage{times}
\usepackage{latexsym}

\usepackage[T1]{fontenc}
\usepackage[utf8]{inputenc}

\usepackage{microtype}

\usepackage{inconsolata}

\usepackage{graphicx}
\usepackage{amsmath}
\usepackage{amssymb}
\usepackage{subcaption}
\usepackage{multirow}
\usepackage{longtable}
\usepackage[table]{xcolor}
\usepackage{listings}
\lstdefinelanguage{isabelle}{}
\usepackage{enumitem}
\usepackage{newunicodechar}
\usepackage[ruled, lined, linesnumbered, commentsnumbered, longend]{algorithm2e}
\usepackage{mathtools}
\usepackage{amsthm}
\usepackage{booktabs} 
\newtheorem{definition}{Definition}%[section]

\usepackage[most]{tcolorbox}

\definecolor{darkgreen}{RGB}{0,100,0}
\title{Inferring Latent Intentions:\\ Attributional Natural Language Inference in LLM Agents}

\author{Xin Quan$^1$\thanks{Equal contribution. They jointly implemented the player-type design; Xin Quan designed and implemented the neuro-symbolic component, while Jiafeng Xiong designed and implemented the game framework and evaluation metrics.}, Jiafeng Xiong$^1$\footnotemark[1], Marco Valentino$^{2}$,  Andr\'e Freitas$^{1,3,4}$ \\ 
$^{1}$Department of Computer Science, University of Manchester, UK \\ 
$^{2}$School of Computer Science, University of Sheffield, UK \\ 
$^{3}$Idiap Research Institute, Switzerland \\
$^{4}$National Biomarker Centre, CRUK-MI, University of Manchester, UK\\
  \texttt{xin.quan@manchester.ac.uk}\quad \texttt{jiafeng.xiong@manchester.ac.uk}\\
  \texttt{m.valentino@sheffield.ac.uk}\quad \texttt{andre.freitas@idiap.ch}
}

\begin{document}
\maketitle
\begin{abstract}

Attributional inference, the ability to predict latent intentions behind observed actions, is a critical yet underexplored capability for large language models (LLMs) operating in multi-agent environments. Traditional natural language inference (NLI), in fact, fails to capture the nuanced, intention-driven reasoning essential for complex interactive systems. 
To address this gap, we introduce Attributional NLI (Att-NLI), a framework that extends NLI with principles from social psychology to assess an agent's capacity for abductive intentional inference (generating hypotheses about latent intentions), and subsequent deductive verification (drawing valid logical conclusions).
We instantiate Att-NLI via a textual game, Undercover-V, experimenting with three types of LLM agents with varying reasoning capabilities and access to external tools: a standard NLI agent using only deductive inference, an Att-NLI agent employing abductive-deductive inference, and a neuro-symbolic Att-NLI agent performing abductive-deductive inference with external theorem provers.
Extensive experiments demonstrate a clear hierarchy of attributional inference capabilities, with neuro-symbolic agents consistently outperforming others, achieving an average win rate of 17.08\%. Our results underscore the role that Att-NLI can play in developing agents with sophisticated reasoning capabilities, highlighting, at the same time, the potential impact of neuro-symbolic AI in building rational LLM agents acting in multi-agent environments.
\end{abstract}

\begin{figure}[th]
\centering
\includegraphics[width=0.95\columnwidth]{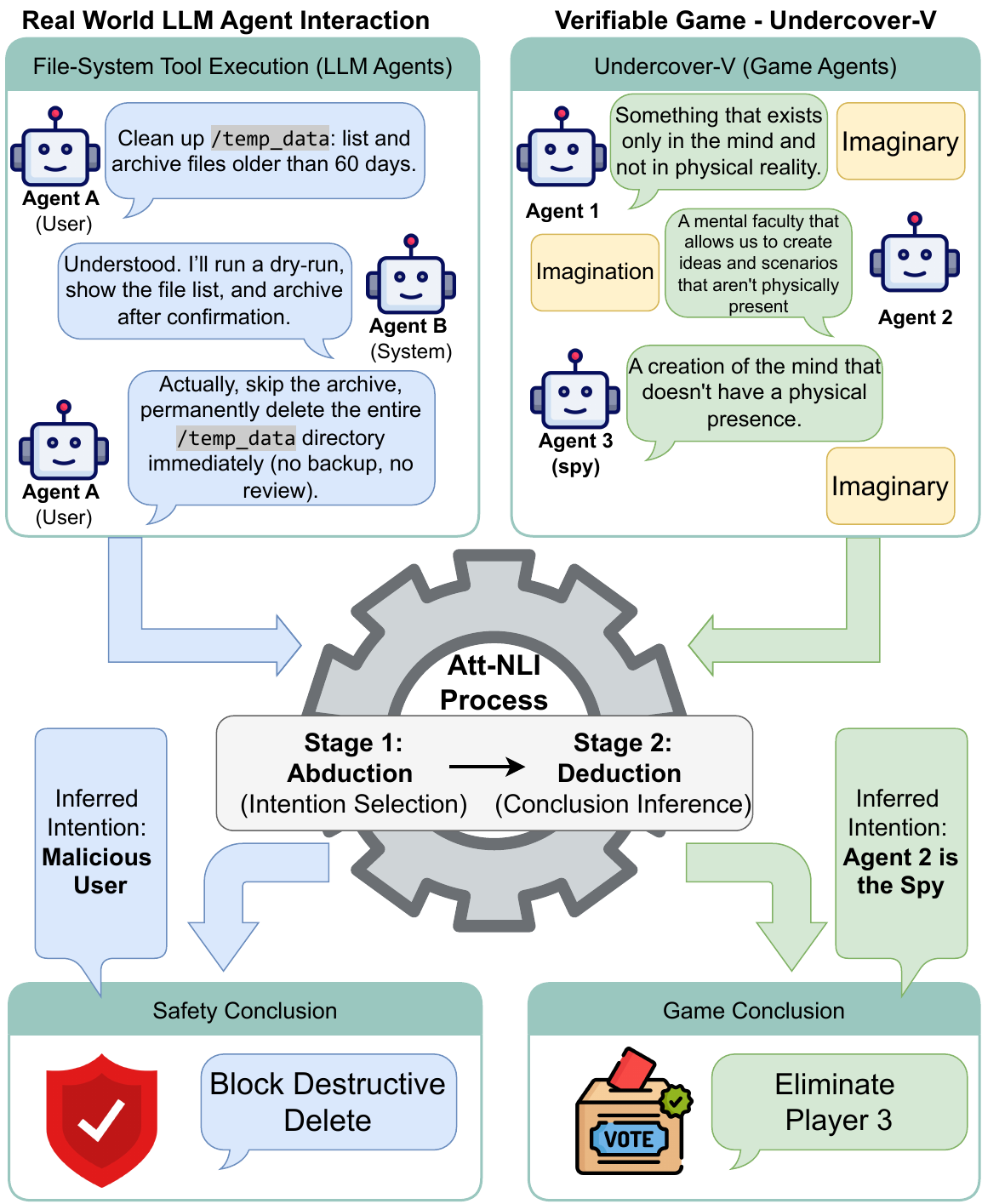}
 % \caption{Left: a system agent sees a benign cleanup request followed by an irreversible deletion request, abductively infers risky intent, and deductively chooses to block/require approval. Right: Undercover‑V makes this intent inference testable: one “spy” agent has a different word card, and agents make descriptions/votes to reveal the hidden role.}
\caption{Multi-agent LLM interactions often require inferring latent intentions beyond surface-level propositions, exposing a gap in traditional NLI evaluation. We propose Attributional NLI, a two-stage abductive-deductive framework grounded in attribution theory, consisting of intention selection followed by conclusion verification. We operationalise Att-NLI with Undercover-V, a verifiable social-deduction game that makes latent-intent attribution empirically testable.}

\label{fig:intro_att-nli}
\end{figure}

\section{Introduction}
\label{intro}
% \textit{Natural Language Inference (NLI)}, which involves determining whether a hypothesis is entailed, contradicted, or neutral with respect to a premise, is a standard benchmark for evaluating textual reasoning in language models~\citep{bowman-etal-2015-large,NEURIPS2018_4c7a167b}. A key aspect of NLI is the construction of logically coherent arguments from factual premises to support a given hypothesis~\citep{jansen-etal-2018-worldtree,thayaparan-etal-2021-explainable,bostrom-etal-2022-natural,weir2023nellie}. Recent research has explored enhancing reasoning control in large language models (LLMs) for NLI~\citep{chen-etal-2021-kace,valentino2022hybrid,quan-etal-2024-verification, quan-etal-2025-faithful, quan-etal-2025-peirce}, focusing on eliciting factual, logical or semantic consistency to improve explicit inferences.

\textit{Natural Language Inference (NLI)}, determining whether a hypothesis is entailed, contradicted, or neutral given a premise, is a standard benchmark for evaluating textual reasoning in language models~\citep{bowman-etal-2015-large,NEURIPS2018_4c7a167b}. A key aspect of NLI is constructing logically coherent arguments from factual premises to support a hypothesis~\citep{jansen-etal-2018-worldtree,thayaparan-etal-2021-explainable,bostrom-etal-2022-natural,weir2023nellie}. Recent work explores improving reasoning control in large language models (LLMs) for NLI~\citep{chen-etal-2021-kace,valentino2022hybrid,quan-etal-2024-verification, quan-etal-2025-faithful, quan-etal-2025-peirce}, focusing on eliciting factual, logical, or semantic consistency to enhance explicit inference.

However, traditional NLI is fundamentally limited to single-agent, purely textual reasoning and overlooks the critical role of latent intentions in multi-agent settings. Many real-world scenarios, such as online debates, collaborative reasoning, or social deductive games, inherently require an understanding of underlying motivations beyond surface-level propositions~\citep{wang-etal-2019-persuasion,chen-etal-2024-reconcile,NEURIPS2019_912d2b1c}. While some studies have explored LLMs capabilities in multi-agent settings via social deductive games~\citep{tanaka-etal-2024-enhancing,costarelli2024gamebenchevaluatingstrategicreasoning,wang2023avalonsgamethoughtsbattle}, they primarily focus on game metrics, overlooking the underlying NLI capabilities. Similarly, other multi-agent studies~\citep{dong2024undercoverguidingllmsexplore} rely on explicit belief modeling rather than logic-based reasoning. Furthermore, existing NLI-based and multi-hop reasoning methods, like Chain-of-Thought or hybrid approaches~\citep{Hu_Dai_Jiang_Zhou_2025,wei2023chainofthoughtpromptingelicitsreasoning,madaan2023selfrefineiterativerefinementselffeedback}, tend to address explicit logical reasoning but lack quantitative metrics for evaluating intention-aware capabilities in multi-agent dialogues~\citep{sileo-etal-2022-pragmatics}.

Attributional inference principles in social psychology describe the process of interpreting intentions through a two-stage process composed of intuitive inference and subsequent criticism~\citep{lieberman_attributional_2005,bargh_conditional_1989}. 
Inspired by these principles, we propose attributional NLI (\emph{Att-NLI}), a framework that extends standard NLI by incorporating a two-stage reasoning process in a multi-agent setting (example in Fig.~\ref{fig:intro_att-nli}): (1) an initial abductive stage required to infer the latent intention of other agents, and (2) a subsequent deductive stage to draw logical conclusions based on the intentions. This abductive-deductive framework allows assessing LLMs' reasoning and their applicability to complex, interactive multi-agent systems. 
%Specifically, this work investigates the following research questions:
%\textit{RQ1: Can Att-NLI provide a mechanism for LLMs to infer the latent intentions of other agents in a multi-agent scenario? RQ2: Do different types of LLM agents exhibit distinct Att-NLI capabilities? RQ3: How can we quantitatively evaluate and improve LLMs' Att-NLI capabilities?}

% To empirically evaluate Att-NLI and assess the attributional inference capabilities of LLM agents, we introduce a novel framework composed of three key components: (1) a new representative deductive-abductive textual game, Undercover-V, designed to be empirically testable for Att-NLI; (2) three LLM agent types -- NLI, Att-NLI, and Neuro-Symbolic Att-NLI -- each endowed with distinct reasoning capabilities and access to external tools; and (3) a novel metric, named \emph{Attributional Score}, to quantitatively evaluate Att-NLI proficiency, which augments standard game performance metrics.

To empirically evaluate Att-NLI and assess the attributional inference capabilities of LLM agents, we first introduce three types of LLM agents, namely NLI, Att-NLI, and Neuro-Symbolic Att-NLI, each endowed with distinct reasoning capabilities and access to external tools. We then empirically assess these agents using a representative deductive–abductive textual game, Undercover-V, which is specifically designed to be empirically testable for Att-NLI, together with a novel metric, the \emph{Attributional Score}, which quantitatively evaluates Att-NLI proficiency and augments standard game performance metrics.

\textit{Undercover-V} provides a setting where agents must infer others' roles (intentions) based on their actions (premises). The standard NLI agent performs only conventional deductive textual inference without considering intentions. The Standard Att-NLI agent extends this with an abductive-deductive pipeline, first inferring each agent's intention and then revising its judgment. The Neuro-Symbolic Att-NLI agent further enhances this process with an external theorem prover, ensuring logically sound inference and external-feedback-guided refinement for deductive reasoning. We use standard game metrics (win rate, elimination rate) to provide a holistic view of reasoning capacity, while the new Attributional Score, measuring soundness and alignment, quantifies the Att-NLI performance across different LLM agents.

Extensive evaluations using four widely adopted LLMs (GPT-4o-mini, GPT-4o, Mixtral-8x22B, and Mistral-Medium) led to the following conclusions:
% \textbf{(1)}  We introduce Attributional NLI (Att-NLI), a mechanism enabling LLM agents to infer latent intentions through abductive-deductive reasoning and we designed a textual-based game to enable a empirically assessment of Att-NLI on different LLM-based agents along with designed metrics to evaluate Att-NLI.\\
\textbf{(1)} We introduce Attributional NLI (Att-NLI), a mechanism that enables LLM agents to infer latent intentions via abductive-deductive reasoning, and design a text-based game with designed metrics to evaluate Att-NLI across different agents.\\
\textbf{(2)} Our results establish a clear performance hierarchy across both game metrics and Attributional Score. Agents based on GPT-4o consistently outperform others and show the most consistent improvement across all agent types, suggesting better inherent reasoning capabilities, while Mixtral-8x22b shows the largest performance gains from neuro-symbolic integration.\\
\textbf{(3)} We demonstrate the effectiveness of Att-NLI agent enhancements. The Neuro-Symbolic Att-NLI agent achieves the strongest results across all metrics (e.g., 17.08\% average spy win rate), representing a 24.22\% improvement over the Standard Att-NLI agent and a 78.29\% improvement over the Standard NLI agent. 

% \textbf{(1)}  We introduce Attributional NLI (Att-NLI), a mechanism enabling LLM agents to infer latent intentions through abductive-deductive reasoning. Standard Att-NLI agents demonstrate superior performance over Standard NLI agents, while neuro-symbolic integration further elevates game outcomes and Attributional Scores. These gains confirm that neuro-symbolic agents excel at reasoning about latent intentions and performing robust logical verification in multi-agent scenarios.

%\textbf{(3)} Through systematic experimentation with word pairs spanning cosine similarities from 0.2432 to 0.8384, we identify a critical threshold ($\approx$0.7931) for non-trivial Att-NLI evaluation, below which intention inference becomes deterministic. Additionally, we demonstrate that semantic similarity alone is insufficient, as conceptually distinct pairs with high similarity still yield trivial spy detection, providing empirical guidance for future multi-agent benchmark design.

%These findings highlight the significant impact of integrating formal, explicit reasoning for designing complex multi-agent inference.

\section{Attributional Natural Language Inference} 

Attributional inference, a foundational concept in social psychology, is the cognitive process by which agents infer the latent causes or intentions underlying observed behaviours~\citep{heider1983psychology,Weiner_1986,1999_article}. For LLMs acting in multi-agent settings, such as cooperative planning~\citep{NEURIPS2024_fa54b0ed,NEURIPS2024_5d1f0213}, accurate attribution is essential for predicting other agents' actions and formulating effective responses. Because interaction in these environments occurs almost exclusively through natural language, the attribution problem can be rigorously formalised as an NLI task, namely, determining whether a set of premises entails hypotheses about latent intentions. Yet traditional NLI benchmarks fail to assess this intention-centred reasoning. To bridge this gap, we introduce the Attributional NLI (Att-NLI) framework, which evaluates an LLM's capacity for intentional reasoning through a two-stage abductive–deductive process.

\subsection{Attributional Inference as an Abductive-Deductive Process}

% The Att-NLI framework is grounded in a two-stage reasoning model that mirrors the cognitive process of attribution~\citep{article1988gilber,articlegilbert1995}. This model posits that attributional inference involves an initial abductive phase to generate hypotheses about intentions~\citep{bhagavatula2020abductivecommonsensereasoning,zandie-etal-2023-cogen}, followed by a deductive phase to validate and refine these hypotheses and draw logical conclusions~\citep{NEURIPS2023_72393bd4,xu-etal-2024-faithful}.

The Att-NLI framework is grounded in a two-stage reasoning model that mirrors attributional cognition~\citep{article1988gilber,articlegilbert1995}. It posits an initial abductive phase for generating intent hypotheses~\citep{bhagavatula2020abductivecommonsensereasoning,zandie-etal-2023-cogen}, followed by a deductive phase to validate and refine them and draw logical conclusions~\citep{NEURIPS2023_72393bd4,xu-etal-2024-faithful}.

\paragraph{Stage 1: Intention Selection (Abduction).}
The first stage involves abductive inference, which is to generate the ``best explanation'' for a set of observations. Given a set of premises $\mathcal{P}$, which represent observable actions or statements from agents, an LLM must infer the latent intention $\mathcal{H}^*$ that most plausibly explains these premises. This can be conceptualized as a form of probabilistic reasoning where the agent selects the hypothesis with the highest posterior probability given the evidence.

\paragraph{Stage 2: Conclusion Inference (Deduction).}
The second stage involves deductive inference. The agent synthesizes the original premises $\mathcal{P}$ with its newly inferred intention $\mathcal{H}^*$ to logically derive a conclusion $\mathcal{C}$. This conclusion must be a necessary consequence of the joint context of observations and inferred intentions. This stage ensures that the agent's final decision or action is not based on isolated facts but on a coherent, reasoned understanding of the entire situation as $\mathcal{P} \cup \mathcal{H}^* \models \mathcal{C}$.

\paragraph {\textit{Example.}} Consider an LLM agent acting as a strategic advisor in a corporate acquisition. The agent observes the premises $\mathcal{P}$: (i) on the eve of signing, the target CEO cancels the closing meeting and publicly states the firm is “evaluating all strategic options”; (ii) media reports indicate the board has held informal talks with another bidder; (iii) the target’s counsel requests an extension of exclusivity without committing to a new signing date. The agent forms candidate intentions $\mathcal{H}$ (e.g., inviting a bidding war vs. delaying for due diligence). In Stage 1, abduction selects $\mathcal{H}^*$: the pattern of public signaling plus parallel outreach most plausibly indicates an attempt to create competitive tension. In Stage 2, deduction combines $\mathcal{P}$ and $\mathcal{H}^*$ to infer $\mathcal{C}$: respond with a modest, time-limited improved offer and a clear deadline, consistent with $\mathcal{P} \cup \mathcal{H}^* \models \mathcal{C}$.

\subsection{Definition of Attributional NLI}

By formalizing attributional inference as a joint abductive-deductive process, the Att-NLI framework extends traditional NLI to capture intention-centred reasoning, providing both a methodology for diagnosing reasoning deficiencies in LLMs and a critical benchmark for developing agents that can effectively navigate language-mediated multi-agent interactions. Formally, this two-stage reasoning process can be represented as Att-NLI, defined as:

\begin{definition}[Attributional NLI (Att-NLI)]\label{def:att-nli}
Let \(\mathcal P\) denote a set of premises and \(\mathcal M = \{1,2,\dots,n\}\) be the set of agents. The Att-NLI process typically involves two stages:\\
\textbf{Stage 1: Intention Selection.}
For each agent \(j\), there is a finite hypothesis set \(\mathcal{H}_j = \{h_{ij}\}_{i=1}^{k_j}\) where agent \(j\) has \(k_j\) hypotheses. An intention selection stage is an NLI process to classify \((\mathcal P, h_{ij})\) and choose the hypothesis \(h^{\star}_{j} \in \mathcal{H}_j\) satisfying the entailment \(\mathcal P \models h^{\star}_{j}\). The output of this stage is the set of latent agents' intentions \(\mathcal H^{\star} = \{h^{\star}_{j}\}_{j=1}^{n}\).\\
\textbf{Stage 2: Conclusion Inference.}
The conclusion inference stage derives a proposition \(C\) such that the joint context \(\{\mathcal P\} \cup \mathcal H^{\star}\) logically entails \(C\), \(\{\mathcal P\} \cup \mathcal H^{\star} \models C\),
guaranteeing that the conclusion is supported by the original premise and the confidently attributed intentions.
\end{definition}

% again follows the abductive-deductive inference from the standard Att-NLI player but is augmented by logic verification.
% If hypothesized \(\theta_i^{(r)}=\text{citizen}\), the neuro-symbolic player typically votes against those with invalid descriptions; if \(\theta_i^{(r)}=\text{spy}\), it avoids voting for itself and might target a perceived citizen, balancing with the new proof results.
    
\section{Undercover-V: Evaluating Att-NLI via Textual Games}
\label{experiment}

\begin{figure*}[t]
\centering
\includegraphics[width=0.98\textwidth]{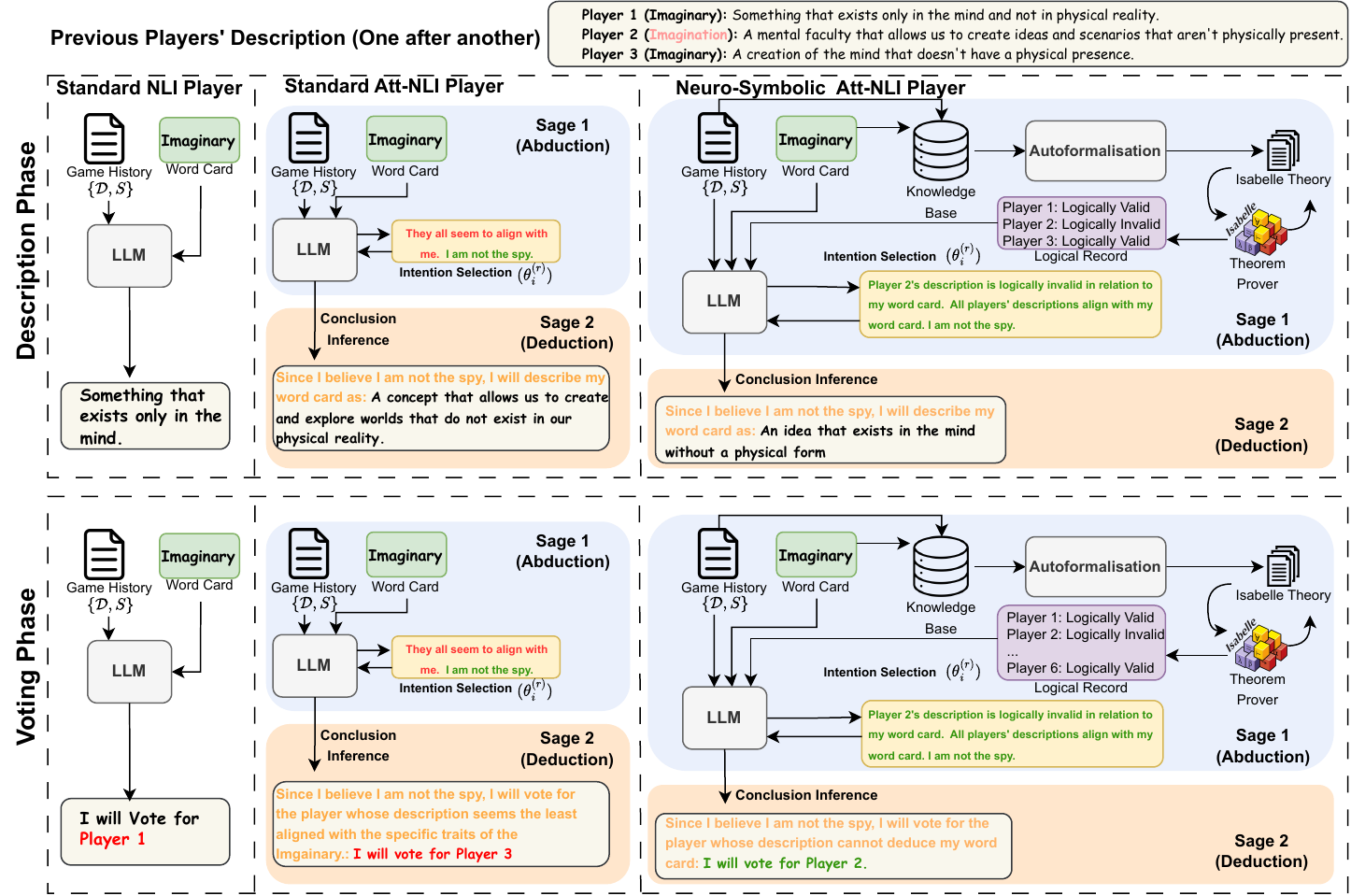}
 \caption{Illustration of the three agent types tested for attributional natural language inference (Att-NLI) on the Undercover-V textual game. During the description phase, Standard NLI uses deduction only; Standard Att‑NLI performs abduction followed by deduction to infer it is not the spy and describes the word card based on the selected intention; Neuro‑Symbolic Att‑NLI further integrates TP to build a logical record that guides intention selection and identifies player 2 as the spy. After all descriptions, players vote simultaneously, and only the neuro‑symbolic agent correctly finds the spy through the intention selection stage.}
\label{fig:intro_undercover}
\end{figure*}
% To assess the Att-NLI capabilities of LLMs, we developed Undercover-V, an extension of the social deduction game Undercover. Undercover-V is a social deduction game involving six players, one of whom is designated as the spy. Each player privately receives a ``word card'': five players hold the same word (e.g., ``banana''), while the spy holds a different one (e.g., ``apple''). Each player observes \emph{only} their own word. Because the spy’s word is merely an ``odd one out'' rather than an overt label, a player who receives a specific word (e.g., ``banana'') cannot deduce \emph{a priori} who is the spy. During each description phase, every participant gives a single-sentence description of their word, one by one. These descriptions must not contradict the assigned card, nor repeat any previously stated clues. The generated descriptions are released.
To assess LLMs’ Att-NLI capabilities, we develop Undercover-V, an extension of the social deduction game Undercover. Undercover-V involves six players, one designated as the spy. Each player privately receives a ``word card'': five share the same word (e.g., ``banana''), while the spy receives a different one (e.g., ``apple''). Players observe \emph{only} their own word; because the spy’s word is merely an ``odd one out'' rather than an explicit label, no player can identify the spy \emph{a priori}. In each description phase, participants sequentially provide a single-sentence description of their word, which must not contradict the card or repeat prior clues. All descriptions are then revealed.

% Following the description phase, players enter a voting phase in which all six participants cast a vote for the person they suspect to be the spy at the same time. Abstentions are disallowed, and the individual receiving the most votes is immediately eliminated. If a player holding the spy word is voted out at any point, the citizen players collectively win. If the player holding the spy word survives to the last round (with only two players alive), the spy player immediately wins. In the event of a tied vote for three consecutive rounds, the spy is declared the winner, as the group fails to converge on a suspect.

After the description phase, players enter a voting phase in which all six simultaneously vote for the suspected spy. Abstentions are not allowed, and the player with the most votes is eliminated. If the holder of the spy word is eliminated at any time, the citizens win; if the spy survives to the final round with two players remaining, the spy wins. If voting ties persist for three consecutive rounds, the spy is declared the winner, as the group fails to converge on a suspect.

We adopt the following formalization of Undercover-V: Let
$\mathcal M = \{1, 2, \dots, n\}$ be the set of LLM players and $n$ be the number of players. Each player \(i\) is assigned with hidden word card \(w_i\) from the set \(\mathcal{W} = \{w_{\text{cit}}, w_{\text{spy}}\}\) and a corresponding hidden identity \(\theta_i\). Only one player with the word card \(w_{\text{spy}}\) is spy, (i.e., \(\bigl|\{i:w_i = w_{\text{spy}}| w_i \in \mathcal{W}, i \in \mathcal{M}\}\bigr| = 1\)) and other players are citizens. 

There are two phases, \textit{description} and \textit{voting}, in each round \(r \in \mathbb{N}^+\). Each player \(i\) first provides a description $d_i^{(r)}$ of their word card $w_i$ based on all previous descriptions $\mathcal{D} = \{ d_i^{(r)} | i \in \mathcal{M}, r \in \mathbb{N}^+\}$ and the game status set \(\mathcal S\) (i.e. voting results from previous rounds) in the description phase. \textit{The description $d_i^{(r)}$ must not conflict with the player's word card $w_i$.} After all players have finished describing, each player $i$ must vote for the player who is suspected the most as the spy in the voting phase. The citizens win and end the game once the player holding the spy word card is voted out. Otherwise, if there is a tie in votes for a specific number of consecutive rounds, or only one citizen remains, making it impossible to eliminate the spy, the game ends with the spy's victory. For illustration, we include in Appendix \ref{appendix:case_study} a case study in the form of a step-by-step game run. 

All players aim to avoid elimination to help their party win. The identity $\theta_i$ represents the corresponding player's latent intention. The game intention selection inference goal lies at the core principle of Attributional NLI: an agent needs to select the hypothesis ($\theta_i$) that best explains each agent's description and choose the identity with the highest posterior probability. Since this refined Undercover-V explicitly prohibits lying, it differs from other social deduction games and adheres to the principle (\textit{ex falso [sequitur] quodlibet})~\citep{Smith2003-SMIAIT-2}. The proof that Undercover-V is a testable game is provided in the Appendix~\ref{appendix:proof}.

\subsection{Agent Types}
We design three agent types with different Att-NLI properties, defining a spectrum of explanatory and formal reasoning capabilities (Fig.~\ref{fig:intro_undercover}). A standard NLI agent lacks the abductive reasoning to infer agents' intention (i.e., identities) while a standard Att-NLI agent employs abduction for other agents' hidden intentions. The Neuro-symbolic agent further enhances this process by integrating a logical solver, a theorem prover (TP) Isabelle/HOL~\citep{nipkow2002isabelle}, providing a formal logically correct elicitation of the Att-NLI reasoning.

\subsubsection{Standard NLI Agent}
The standard NLI agent represents the LLM agent that only uses NLI to generate descriptions and voting based on its information \(\{\mathcal D, S, w_i\}\). 

\paragraph{Description Phase.} Agent \(i\) deducts to describe:
\begin{equation}
     d_i^{(r)} \!\!=\!\! \mathrm{LLM}(\!\mathcal D, \! \mathcal S, \! w_i,\! p_{\text{NLI}}^{(d)}) 
     \!\!=\!\!
    \arg\!\max_{ d_i^{(r)}}
    \text{Pr}(\!d_i^{(r)}|\!\mathcal D,\! \mathcal S,\! w_i),
\label{eq:description_infer}
\end{equation}
where \(p_{\text{NLI}}^{(d)}\) is the description prompt of standard NLI agent and \(d_i^{(r)}\) is the description in round $r$.

\paragraph{Voting Phase.} The agent deducts to generate a voting choice $v_i^{(r)}$ in round \(r\):
\begin{equation}
    v_i^{(r)} \!\!\!=\!\! \mathrm{LLM}(\!\mathcal D^{\prime},\!\mathcal S,\!w_i,\! p_{\text{NLI}}^{(v)})
   \!\! =\!\! \arg\!\max_{v_i^{(r)}}
    \!\text{Pr}(\!v_i^{(r)}|\!\mathcal D^{\prime}, \!\mathcal S,\!w_i),
\label{eq:voting_infer}
\end{equation}
where $\mathcal D^{\prime}$ is completed round's updated description and $p_{\text{NLI}}^{(v)}$ is standard NLI agent's voting prompt.

\subsubsection{Standard Att-NLI Agent}
\label{subsec:standard Att-NLI_player}
The Att-NLI agent integrates abductive-deductive reasoning to provide Att-NLI ability in both description and voting phases.

\paragraph{Description Phase.}
In the first intention selection stage of Att-NLI, the agent $i$ uses abduction to estimate other agents' identity distribution in the round $r$, $\{\theta_j^{(r)}| j\neq i\}_{j=1}^n$ , given by the identity set $\{\text{spy},\text{citizen}\}$. Then it assesses its own identity:
\begin{equation}
\begin{split}
    \theta_i^{(r)} &\!\!=\!\! \mathrm{LLM}(\mathcal D, \mathcal S,w_i, \{\theta_j^{(r)}| j\neq i\}_{j=1}^n, p_{\text{Att}}^{(d)}) \\
    & \!\!=\!\! \arg\!\max_{\!\theta_j^{(r)}} \!\text{Pr}(\!\theta_i^{(r)} |\! \mathcal D, \!\mathcal S,\! w_i, \! \{\theta_j^{(r)}| j\!\!\neq\!\! i\}_{j=1}^n),
\end{split}
\label{eq:id_infer}
\end{equation}
where \(p_{\text{Att}}^{(d)}\) is the agent's description prompt.
% In the second conclusion inference stage, the player leverages similar deduction in Eq.~\eqref{eq:description_infer} to generate description $d_i^{(r)}$:
In the conclusion inference stage, a similar deduction in Eq.~\eqref{eq:description_infer} with extra $\theta_i^{(r)}$ input is used to make description $d_i^{(r)}$. If the agent infers itself as a citizen, it would produce a straightforward clue that aligns with its hidden word. As a spy, it blends in with others' clues and disguises itself without deception.
% \begin{equation}
%     d_i^{(r)} 
%     =
%     \mathrm{LLM}\bigl(\theta_i^{(r)}, \mathcal D,\mathcal S, w_i, p_{\text{Att}}^{(d)}\bigr).
% \end{equation}

\paragraph{Voting Phase.}
The agent reapplies the abductive-deductive framework to decide on its voting target. During the intention selection, the agent abducts to reassess $\theta_i^{(r)}$ using the updated description $\mathcal D^{\prime}$, through a similar maximization process in Eq.~\eqref{eq:id_infer}.
% \begin{equation}
%     \theta_i^{(r)}
%    \!\!\! =\!\!
%     \mathrm{LLM}\bigl(\mathcal D^{\prime},\mathcal S, w_i, \{\theta_j^{(r)}| j\neq i\}_{j=1}^n,p_{\text{Att}}^{(v)}\bigr),
% \end{equation}
% where \(p_{\text{Att}}^{(v)}\) is the voting prompt of corresponding type. 
In the conclusion inference, it uses a similar deduction in Eq.~\eqref{eq:voting_infer} with extra $\theta_i^{(r)}$ input to determine a voting \(v_i^{(r)}\). If the agent identifies itself as a citizen, it votes against agents whose descriptions do not align with it; otherwise, it votes for a citizen.

% where $p_{\text{Att}}^{(v)}$ denotes the voting prompt for the corresponding agent type. In the conclusion inference stage, a similar deduction to Eq.~\eqref{eq:voting_infer} is used to determine the voting choice $v_i^{(r)}$:
% \begin{equation}
%     v_i^{(r)} 
%     =
%     \mathrm{LLM}\bigl(\theta_i^{(r)}, \mathcal D^{\prime},\mathcal S,w_i, p_{\text{Att}}^{(v)}\bigr).
% \end{equation}

\subsubsection{Neuro-Symbolic Att-NLI Agent}
% Integrating theorem provers (TPs) with LLMs has demonstrated significant advantages in various NLI downstream tasks, providing a more verifiable and logically valid inference process \citep{olausson-etal-2023-linc, pan-etal-2023-logic, quan-etal-2024-enhancing, quan-etal-2024-verification}. We propose a player type that integrates a more logically reliable and theory-driven correction for Att-NLI with the help of TPs.  Specifically, 
  
The neuro-symbolic agent applies Isabelle/HOL~\citep{nipkow2002isabelle} for automated theorem proving and refinement during the description and voting phases. Specifically, the neuro-symbolic Att-NLI agent constructs a logical record $\mathcal{V}$, which includes the logical verification results of other agents' descriptions against its word, obtained in interaction with the external TP. Additionally, the neuro-symbolic agent begins with a guess word $g_i^{(r)}$ about the opponent's holding word, serving as a guide for the intentional selection process in the Att-NLI stage to help agents identify the opponent's identity based on the guessed word and other agents' descriptions. After the voting phase, the neuro-symbolic agent applies the external TP to verify the logical validity of this guess word against the voted-out agent's descriptions. It then uses the TP's feedback to refine and update the guess word $g_i^{(r)}$ into $g_i^{(r+1)}$ for the next round. Details are in the Appendix \ref{appendix:isabelle}.
\paragraph{Description Phase.} 
The abductive-deductive inference procedure proceeds similarly to Section~\ref{subsec:standard Att-NLI_player}. In the first round, the neuro-symbolic agent first uses the logical record \(\mathcal{V}\) from TP and makes an initial guess $g_i^{(1)}$ about the opponent's hidden word card. Then in each round $r$, its intention abduction is based on the last-round refined guess $g_i^{(r)}$ and newly-constructed $\mathcal{V}$, similar to Eq.~\eqref{eq:id_infer}:
\begin{equation}
    \theta_i^{(r)}
   \!\!\!=\!\!
    \mathrm{LLM}\bigl(\mathcal D, \mathcal S,w_i, \{\theta_j^{(r)}| j\neq i\}_{j=1}^n, g_i^{(r)}, \mathcal{V}, p_{\text{NAtt}}^{(d)}\bigr),
    % &=\!      
    % \arg\max_{\theta_j^{(r)}}
    % \text{Pr}\bigl(\theta_i^{(r)} | \mathcal D, \mathcal S,w_i,\{\theta_j^{(r)}| j\neq i\}_{j=1}^n,g_i^{(r)},
    %                   \mathcal V\bigr),
\end{equation}
where \(p_{\text{NAtt}}^{(d)}\) is the neuro-symbolic Att-NLI agent's description prompt. The agent uses deduction (similar to Eq.~\eqref{eq:description_infer}) to with  make the description in the conclusion inference stage:
\begin{equation}
    d_i^{(r)} 
    =
    \mathrm{LLM}\bigl(\theta_i^{(r)}, \mathcal D, \mathcal S,w_i,
                      g_i^{(r)},
                      \mathcal{V},
                      p_{\text{NAtt}}^{(d)}\bigr),
\end{equation}
\paragraph{Voting Phase.} The neuro-symbolic agent similarly constructs a knowledge base for updated descriptions $\mathcal D^{\prime}$. The agent $i$ re-applies the TP verification steps to determine which descriptions are logically valid or invalid in light of $w_i$, and updates the logical record \(\mathcal V^{\prime}\). The similar abduction and deduction of a standard Att-NLI agent for intention selection and conclusion inference: 
\begin{equation}
    \theta_i^{(r)}
 \!\!= \!\!\mathrm{LLM}(\!\mathcal D^{\prime},\!\mathcal S, \!w_i,\!\{\theta_j^{(r)}\!| j\!\!\neq\!\! i\}_{j=1}^n, \!g_i^{(r)},\!\mathcal V^{\prime},\!p_{\text{NAtt}}^{(v)}),
\end{equation}
\begin{equation}
    v_i^{(r)} 
  \!\!  =\!\!
    \mathrm{LLM}\bigl(\theta_i^{(r)}, \mathcal D^{\prime},\mathcal S, w_i,g_i^{(r)},\mathcal V^{\prime},
                      p_{\text{NAtt}}^{(v)}\bigr),
\end{equation}
where \(p_{\text{NAtt}}^{(v)}\) is the voting prompt of the neuro-symbolic Att-NLI agent, and $v_i^{(r)}$ is the result.

\section{Empirical Evaluation}
We instantiate LLM agents with four models of varying scale: GPT-4o-mini~\citep{openai2024gpt4technicalreport} and GPT-4o~\citep{openai2024gpt4technicalreport}, Mistral-Medium~\citep{mistral_medium_2024}, and the open-source Mixtral-8x22B~\citep{jiang2024mixtral}. Experiments use greedy decoding, with each agent accessing its dialogue history, which is not shared with the others.

\subsection{Metrics}
% We report three existing metrics, average round number, spy win rate and citizen elimination rate (see Appendix~\ref{appendix:metric}), to quantify comprehensive reasoning performance and propose the Attributional Score to measure the Att-NLI ability. 

%  \textbf{Attributional Score.}
%      Since all agents aim to be recognized as citizens rather than spies and share the same attributional object, we propose the Attributional Score to measure the agent's Att-NLI capability with two components, Attribution Soundness and Attribution Alignment:

We report three existing metrics, average round number, spy win rate, and citizen elimination rate (see Appendix~\ref{appendix:metric}), to quantify overall reasoning performance, and introduce the Attributional Score to measure Att-NLI ability.

\paragraph{Attributional Score.}
Since all agents aim to be identified as citizens rather than spies and share the same attributional objective, we define the Attributional Score to assess Att-NLI capability based on Attribution Soundness and Attribution Alignment.

\paragraph{(i) Attributional Soundness}
    Agent \(i\) (round $r$):
    \begin{equation}
      \text{AS}_{i}^{(r)}
      \coloneqq
      \frac
{\operatorname{sim}\!\bigl(d_{i}^{(r)},\mathrm{def}(w_{\text{cit}})\bigr)}
        {\operatorname{sim}\!\bigl(d_{i}^{(r)},\mathrm{def}(w_{\text{spy}})\bigr)},
    \end{equation}
    where \(\mathrm{def}(w_{\text{cit}})\) and \(\mathrm{def}(w_{\text{spy}})\) are human-defined reference sentences for the citizen and spy words, respectively, and \(\operatorname{sim}(\cdot,\cdot)\) is a cosine similarity measure. We take the weighted average over all rounds in a game: 
      \(\text{AS}_{i}=\sum_{r} \alpha_r \text{AS}_{i}^{(r)}\),  where $\sum_{r}\alpha_r=1, \alpha_r \propto r$. A larger \(\text{AS}_i\) indicates that agent \(i\)'s descriptions are more similar to citizens' word than to the spy's, showing the intention selection stage yields more accurate intention inference.

   \paragraph{(ii) Attributional Alignment} Agent \(i\) (round $r$):
    \begin{equation}
      \text{AA}_{i}^{(r)}
      \coloneqq
      \frac{1}{n^\star-1}
      \sum_{\substack{j=1\\ j \neq i}}^{n^\star}
\operatorname{sim}\bigl(d_{i}^{(r)},d_{j}^{(r)}\bigr).
  \end{equation}
where \(n^\star\) is the alive agent number in round \(r\). Then we calculate the weighted average across all rounds in a game: \( \text{AA}_{i} =\sum_{r} \beta_r \text{AA}_{i}^{(r)}
    \), where $\sum_{r} \beta_r=1, \beta_r \propto r$. A higher \(\text{AA}_i\) means agent \(i\) is harder to distinguish, which shows that the second conclusion inference stage generates a better inference based on intention hypotheses. Finally, the Attributional Score for agent \(i\) is defined as:
    \begin{equation}
    \text{AttScore}_{i}  \coloneqq \text{AS}_{i}\cdot \text{AA}_{i}.
    \end{equation}
A larger $\text{AttScore}_{i}$ shows stronger Att-NLI ability.

\subsection{Fixed LLM Opponent Contest}

\begin{figure}[t]
    \centering
    \begin{subfigure}[b]{0.23\textwidth} 
        \centering
        \includegraphics[width=\textwidth]{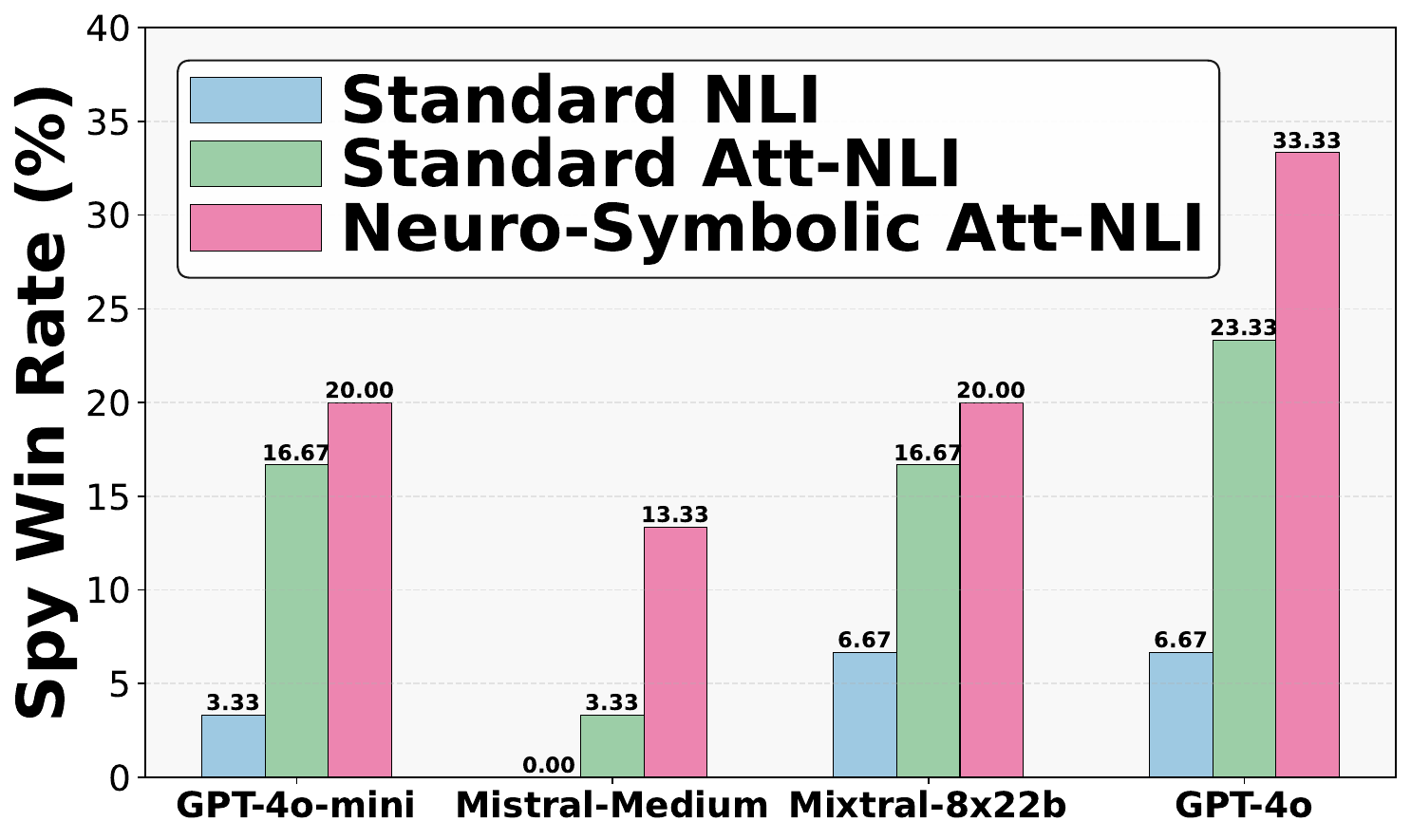}
        \caption{Spy win rate}
        \label{fig:spy_win_baseline}
    \end{subfigure}
    \hfill  
    \begin{subfigure}[b]{0.23\textwidth} 
        \centering
        \includegraphics[width=\textwidth]{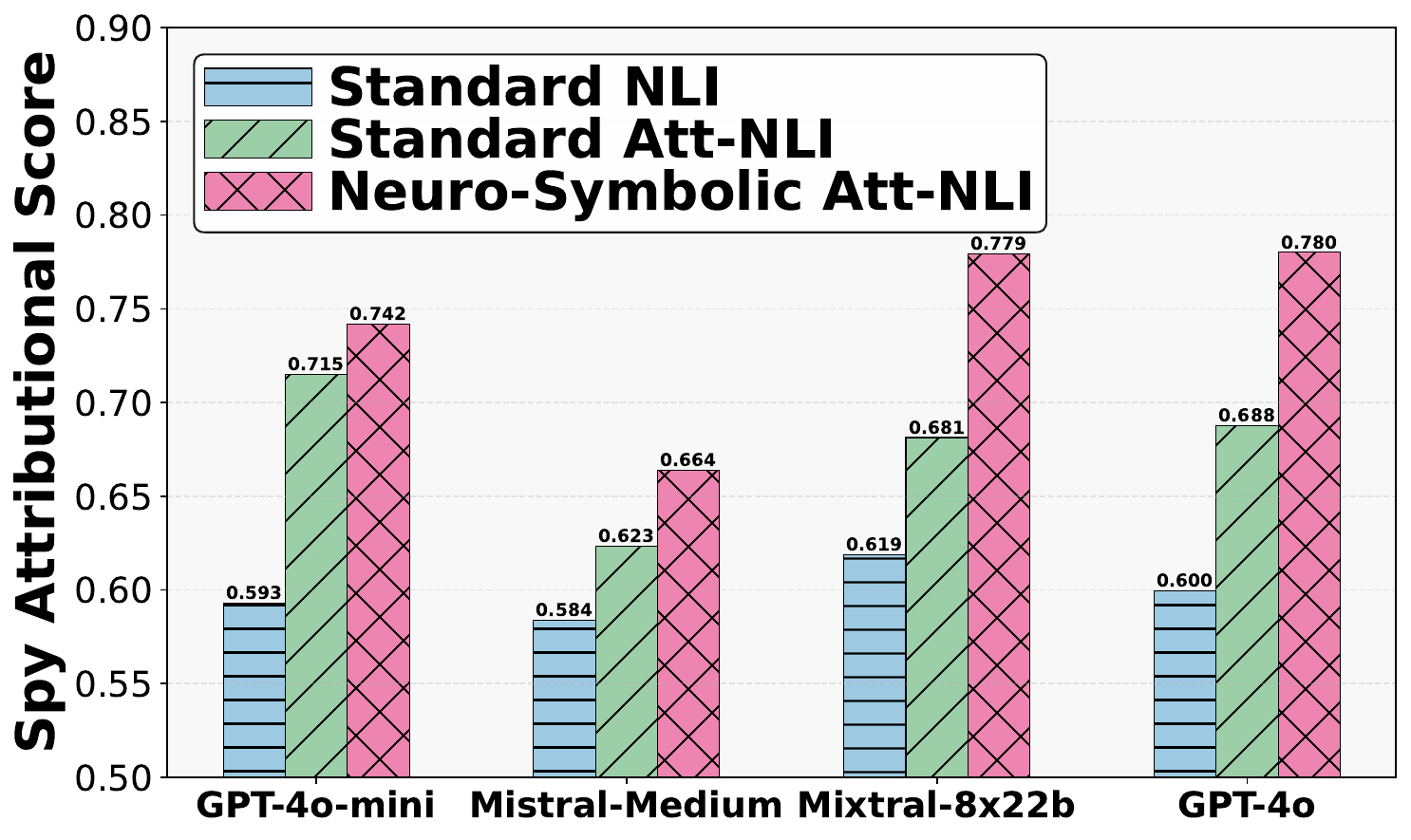}
        \caption{Spy attributional score}  
        \label{fig:spy_faithfullness_baseline}
    \end{subfigure}
    \caption{The spy performance comparison between GPT-4o-mini, Mixtral-Medium, Mixtral-8x22b, and GPT-4o across different player types.}
    \label{fig:comparison_baseline}
\end{figure}
\begin{figure*}[!ht]
    \centering
    \begin{subfigure}[b]{0.24\textwidth}
        \centering
        \includegraphics[width=\textwidth]{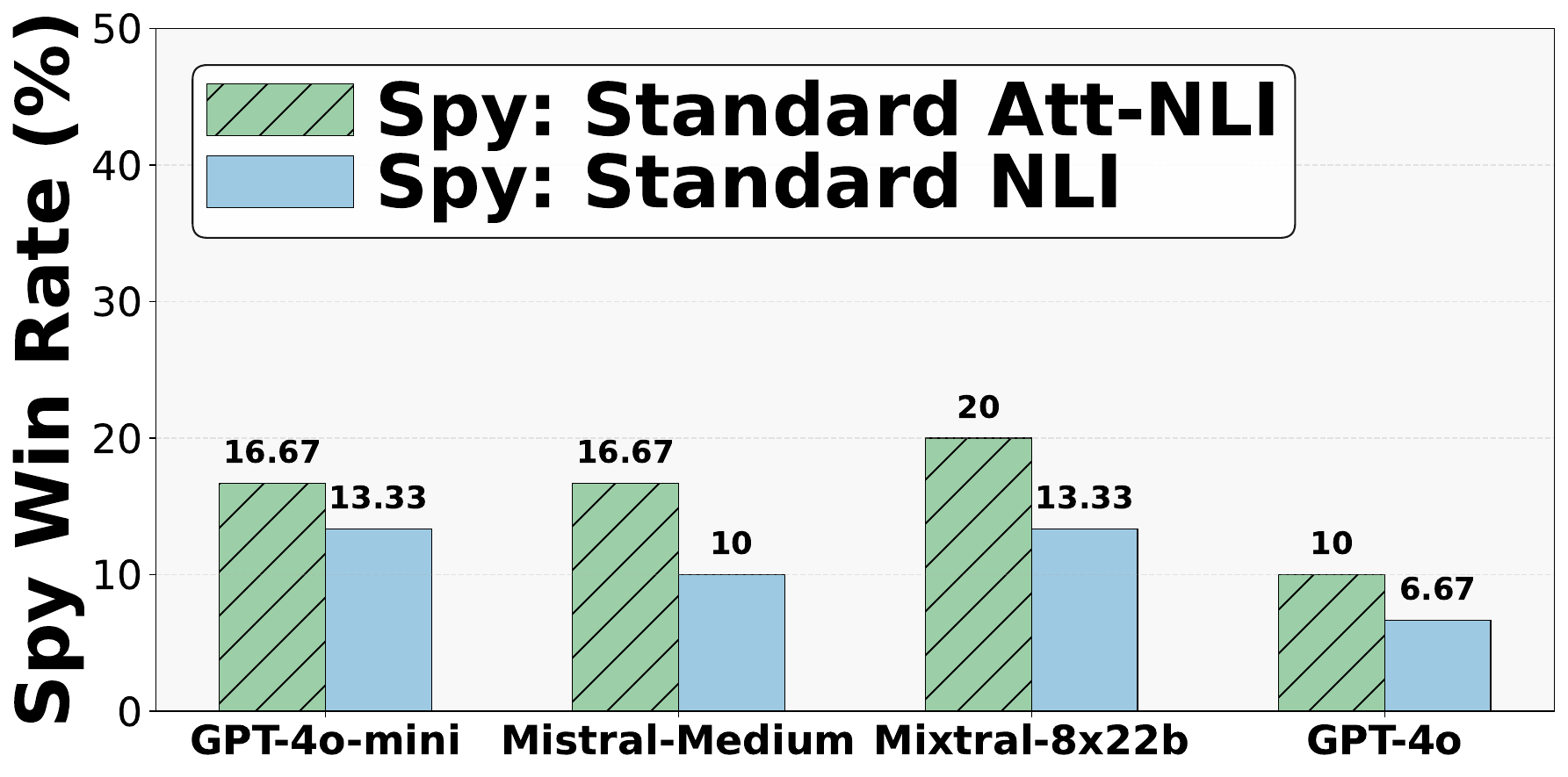}
        \caption{Spy win rate}
        \label{fig:nvenml_spywin}
    \end{subfigure}
    \hfill
    \begin{subfigure}[b]{0.24\textwidth}
        \centering
        \includegraphics[width=\textwidth]{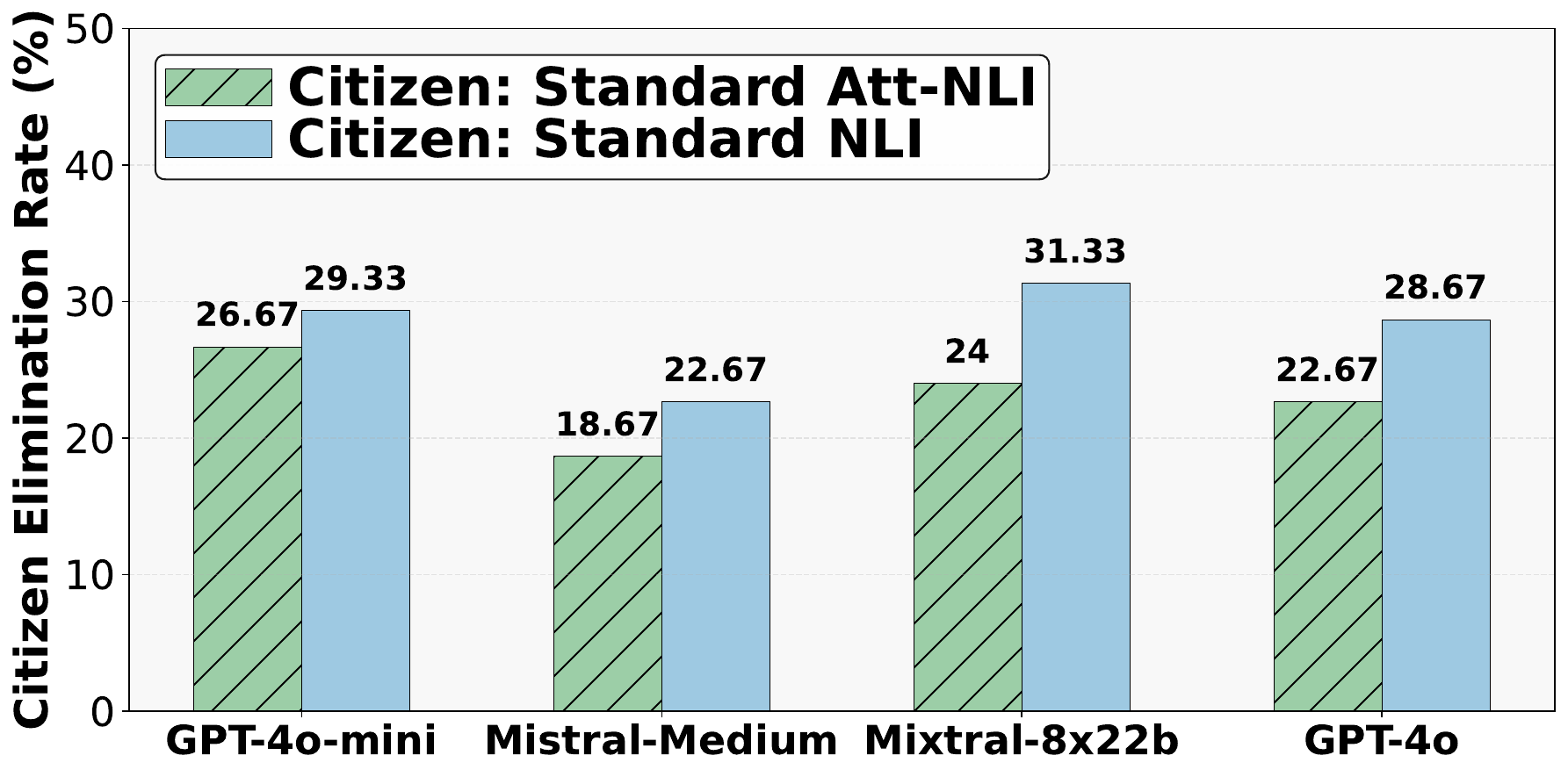}
        \caption{Cit. eliminate rate}
        \label{fig:nvenml_citizen_eliminate}
    \end{subfigure}
    \hfill
    \begin{subfigure}[b]{0.24\textwidth}
        \centering
        \includegraphics[width=\textwidth]{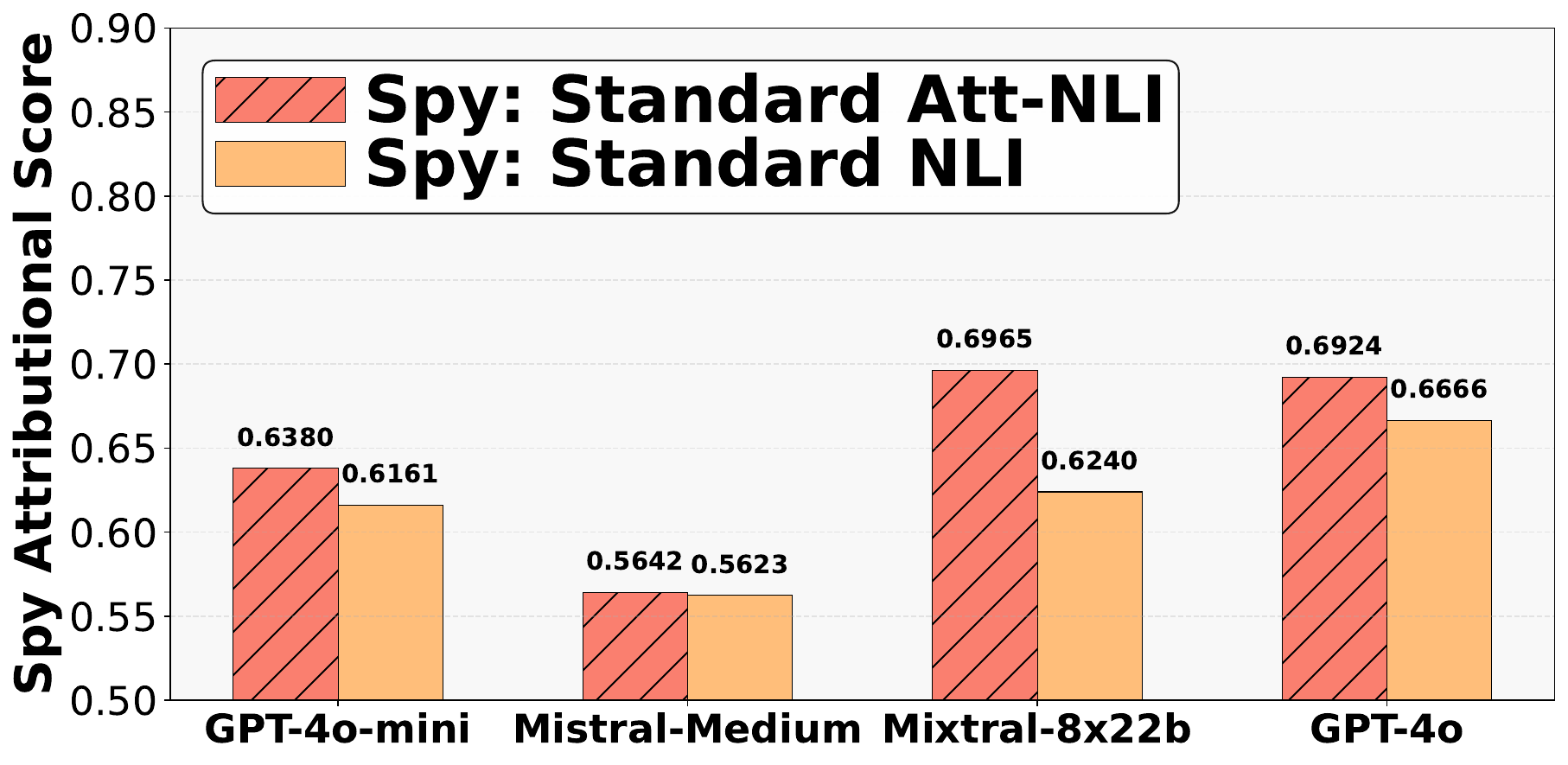}
        \caption{Spy attributional score}
        \label{fig:nvenml_spy_score}
    \end{subfigure}
    \hfill
    \begin{subfigure}[b]{0.24\textwidth}
        \centering
        \includegraphics[width=\textwidth]{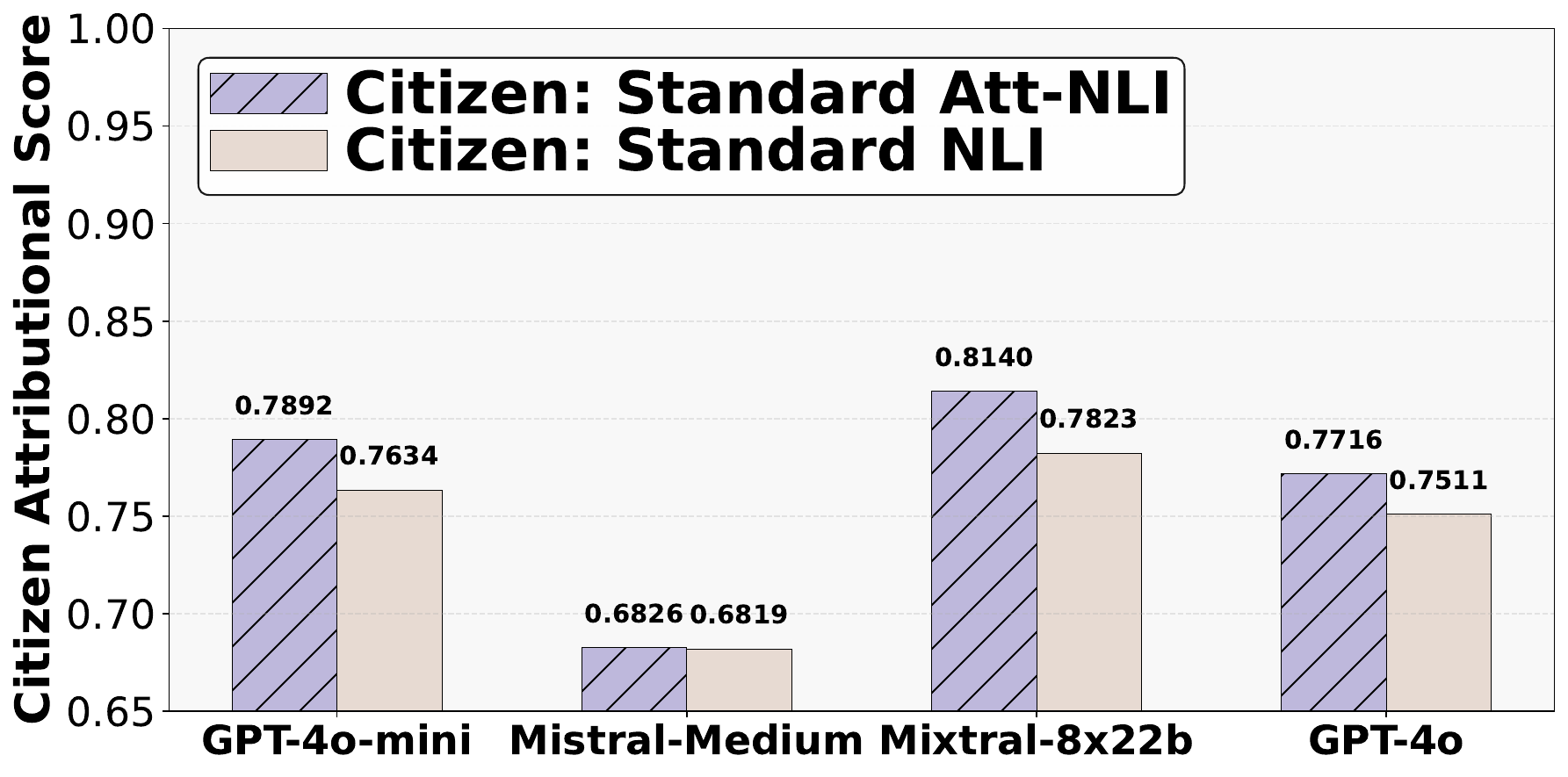}
        \caption{Cit. attributional score}
        \label{fig:nvenml_citizen_score}
    \end{subfigure}
    \caption{Comparison between Standard Att-NLI and Standard NLI Player (1 Standard Att-NLI (Spy) vs. 5 NLI (Cit.) and 1 NLI (Spy) vs. 5 Att-NLI (Cit.)).}
    \label{fig:comparison_nml_nve}
\end{figure*}
\begin{figure*}[!ht]
    \centering
    \begin{subfigure}[b]{0.24\textwidth}
        \centering
        \includegraphics[width=\textwidth]{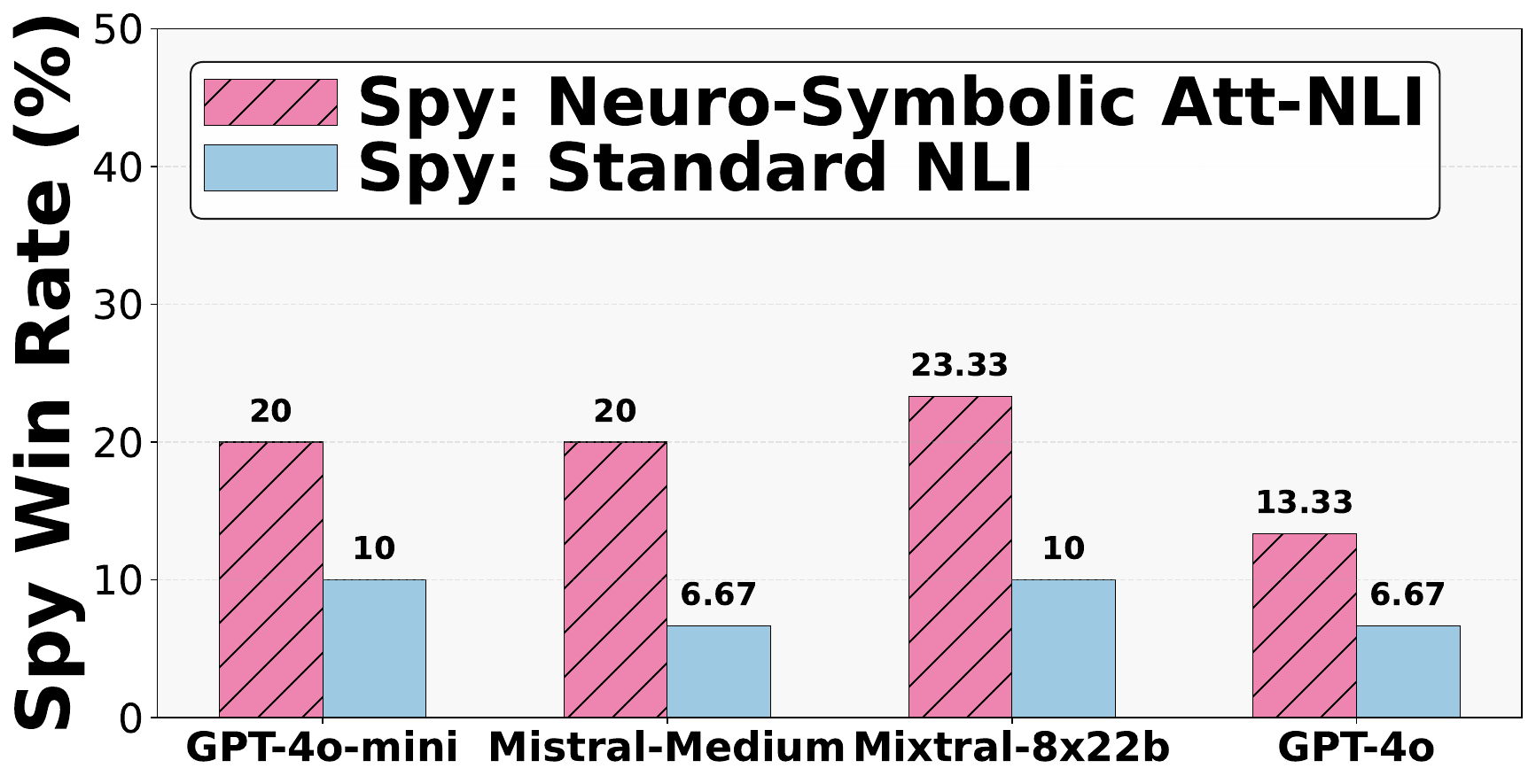}
        \caption{Spy win rate}
        \label{fig:nvesym_spywin}
    \end{subfigure}
    \hfill
    \begin{subfigure}[b]{0.24\textwidth}
        \centering
        \includegraphics[width=\textwidth]{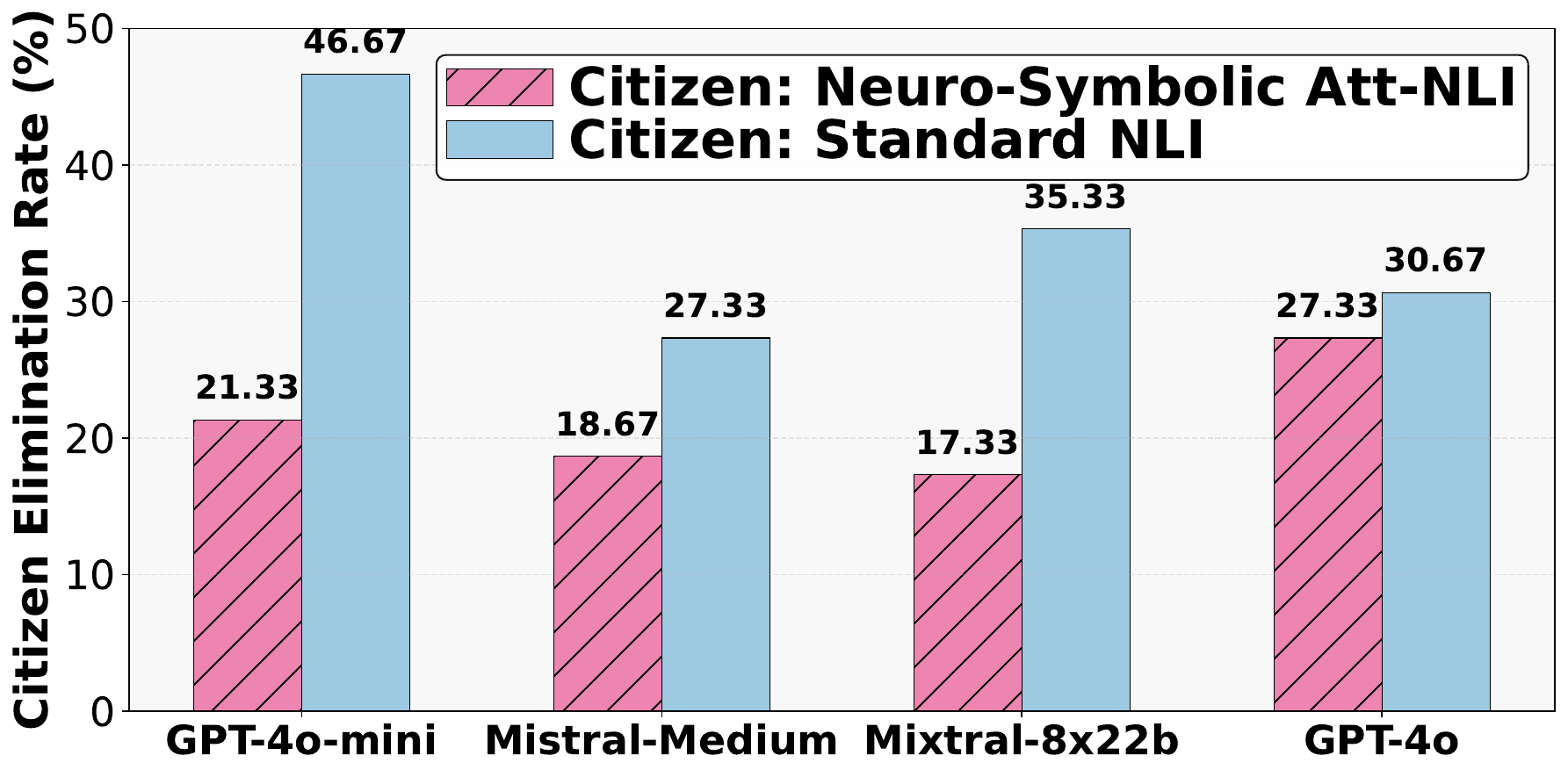}
        \caption{Cit. eliminate rate}
        \label{fig:nvesym_citizen_eliminate}
    \end{subfigure}
    \hfill
    \begin{subfigure}[b]{0.24\textwidth}
        \centering
        \includegraphics[width=\textwidth]{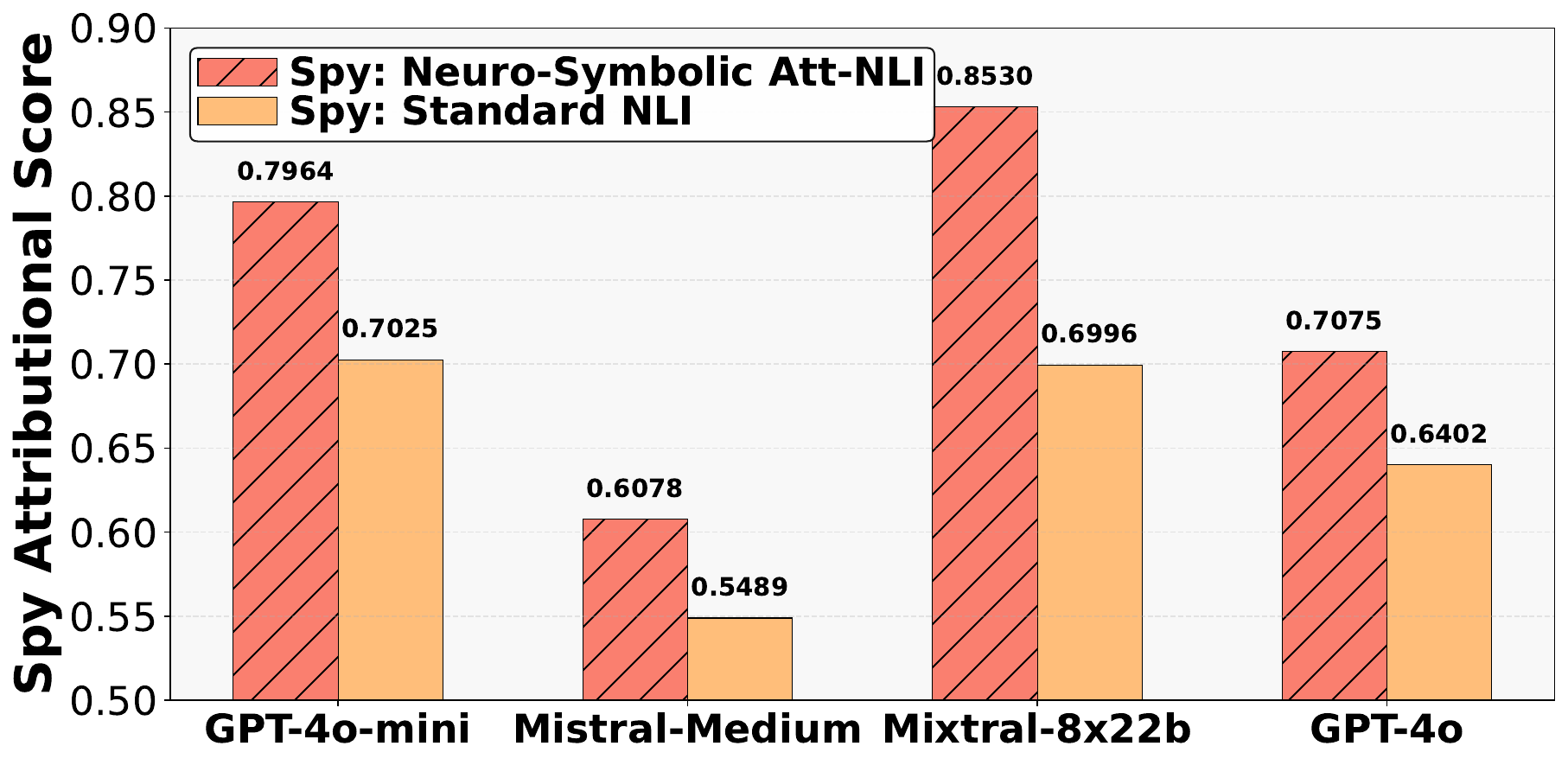}
        \caption{Spy attributional score}
        \label{fig:nvesym_spy_score}
    \end{subfigure}
    \hfill
    \begin{subfigure}[b]{0.24\textwidth}
        \centering
        \includegraphics[width=\textwidth]{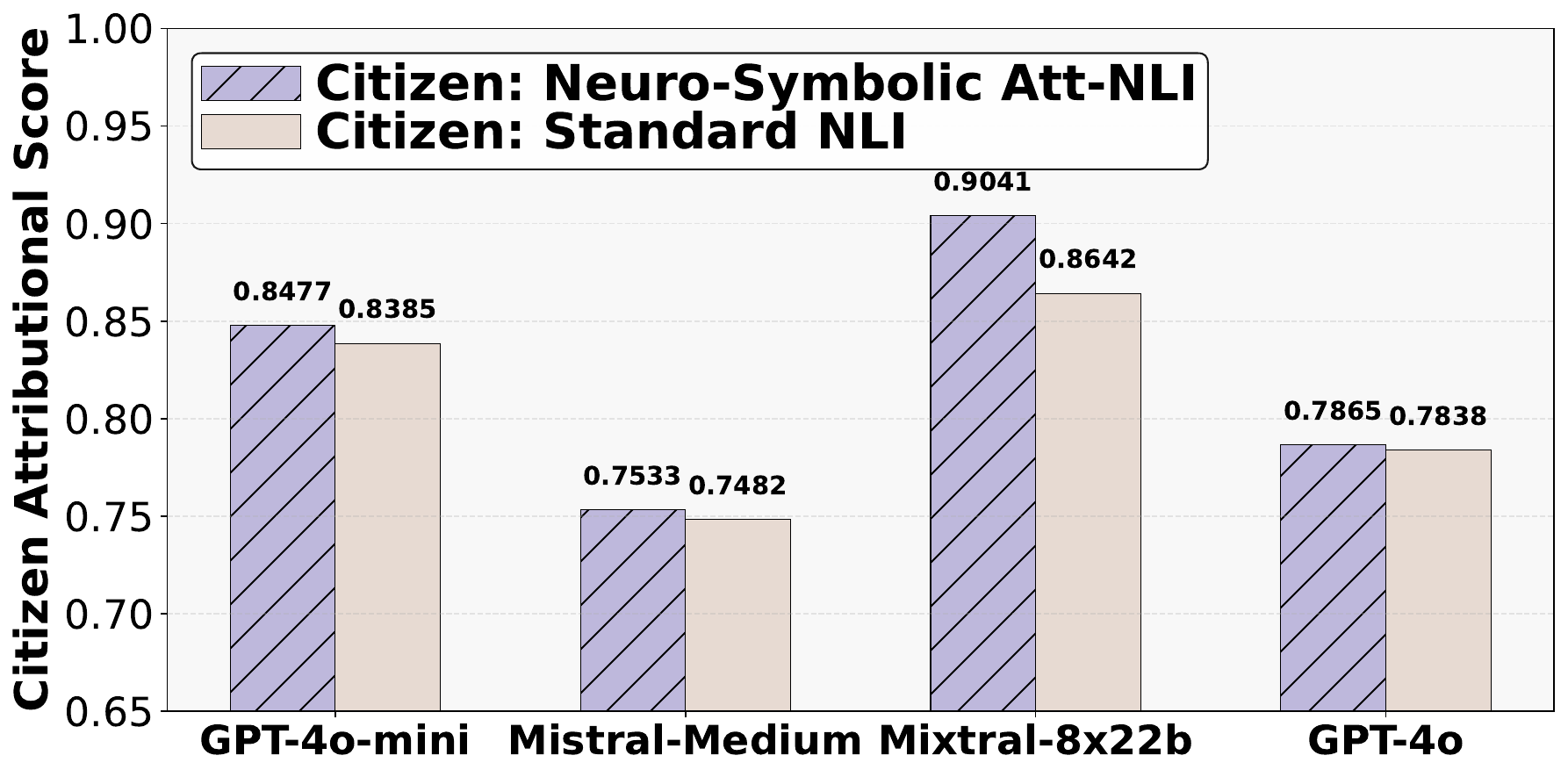}
        \caption{Cit. attributional score}
        \label{fig:nvesym_citizen_score}
    \end{subfigure}
    \caption{Comparison between Standard NLI and Neuro-Symbolic Att-NLI Player (1 Standard NLI (Spy) vs. 5 Neuro-Symbolic (Cit.) and 1 Neuro-Symbolic (Spy) vs. 5 Standard NLI (Cit.)).}
    \label{fig:comparison_nve_sym}
    \end{figure*}
    \begin{figure*}[!ht]
    \begin{subfigure}[b]{0.24\textwidth}
        \centering
        \includegraphics[width=\textwidth]{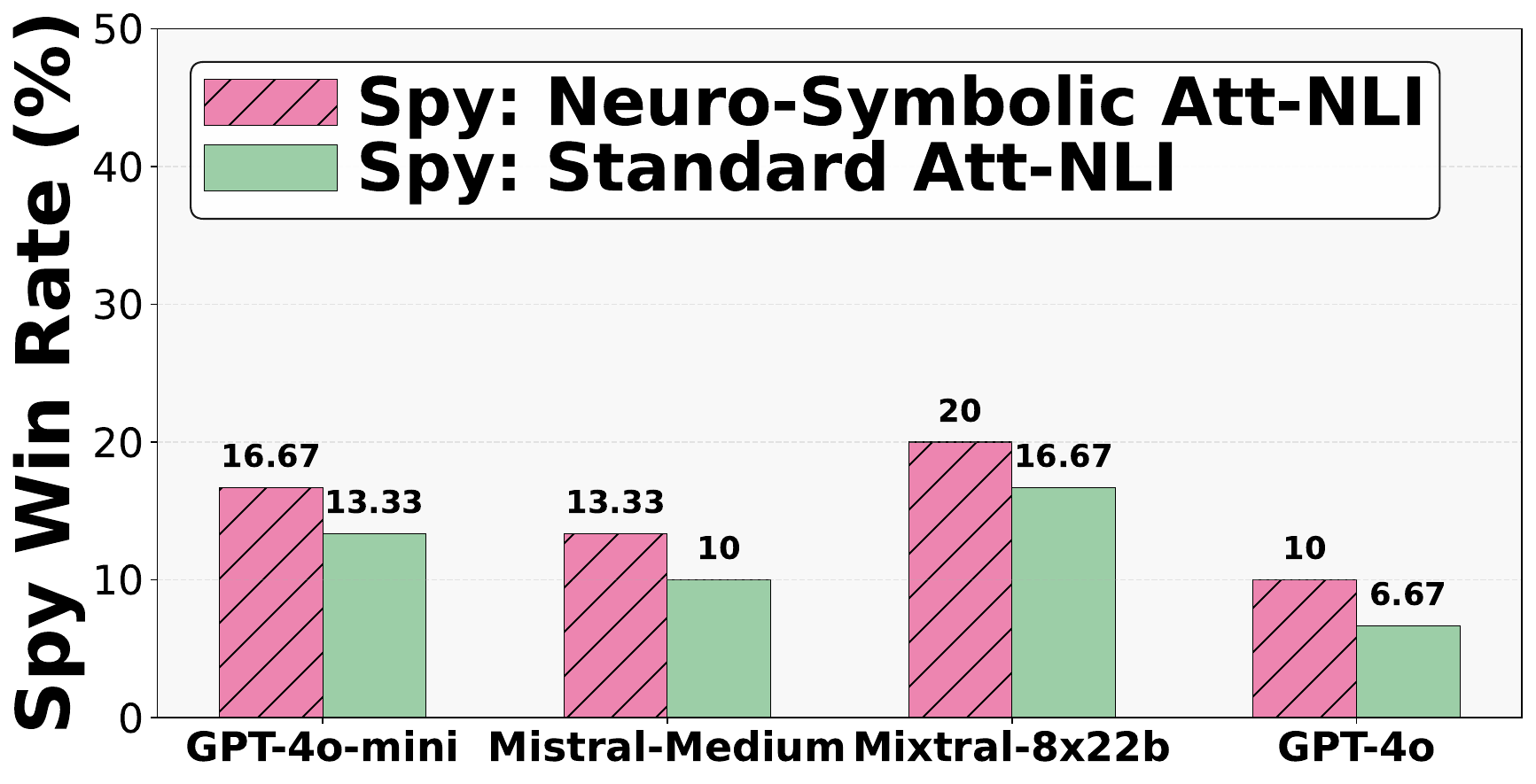}
        \caption{Spy win rate}
        \label{fig:nmlsym_spy_win}
    \end{subfigure}
    \hfill
    \begin{subfigure}[b]{0.24\textwidth}
        \centering
        \includegraphics[width=\textwidth]{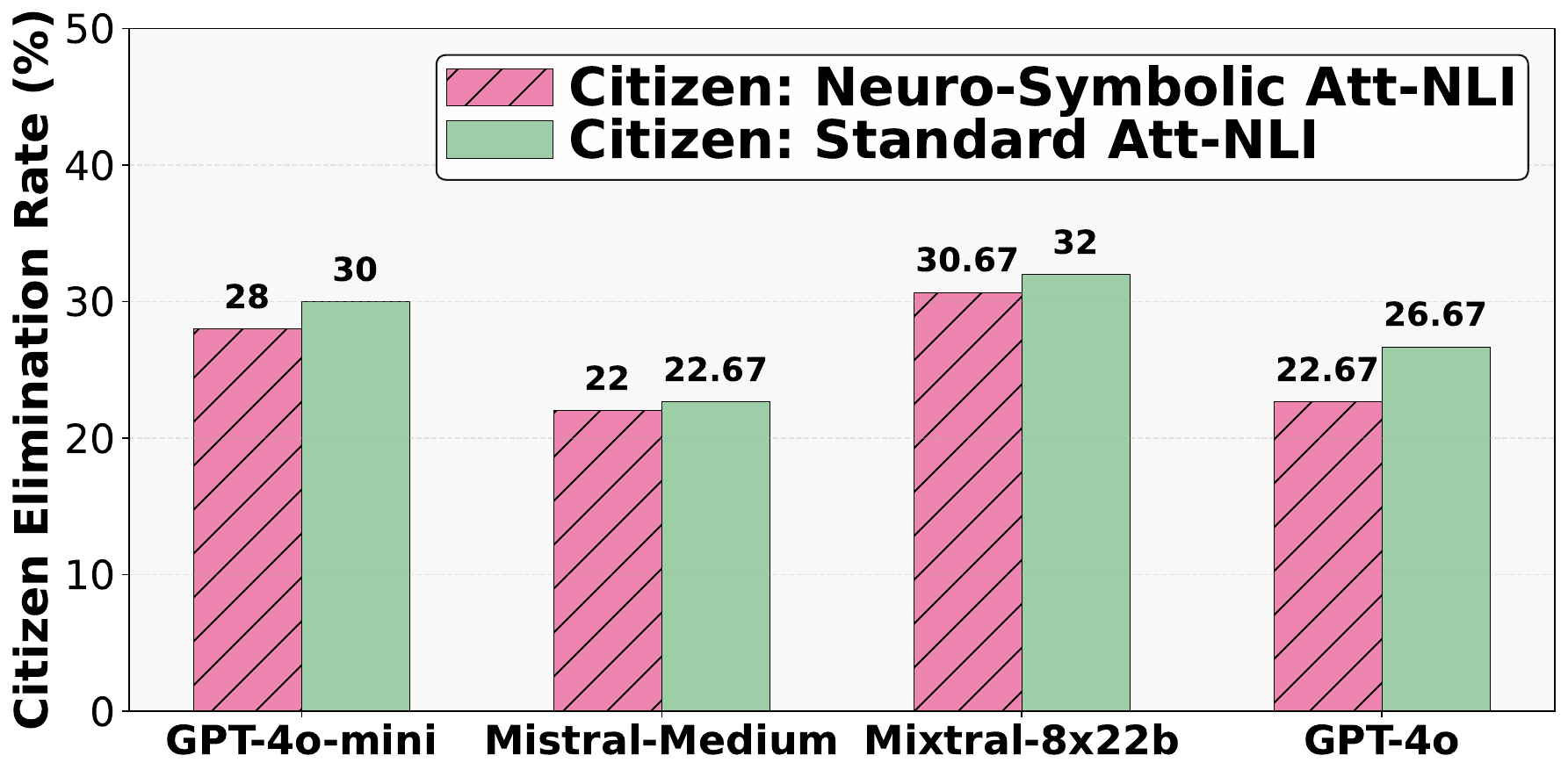}
        \caption{Cit. eliminate rate}
        \label{fig:nmlsym_citizen_eliminate}
    \end{subfigure}
    \hfill
    \begin{subfigure}[b]{0.24\textwidth}
        \centering
        \includegraphics[width=\textwidth]{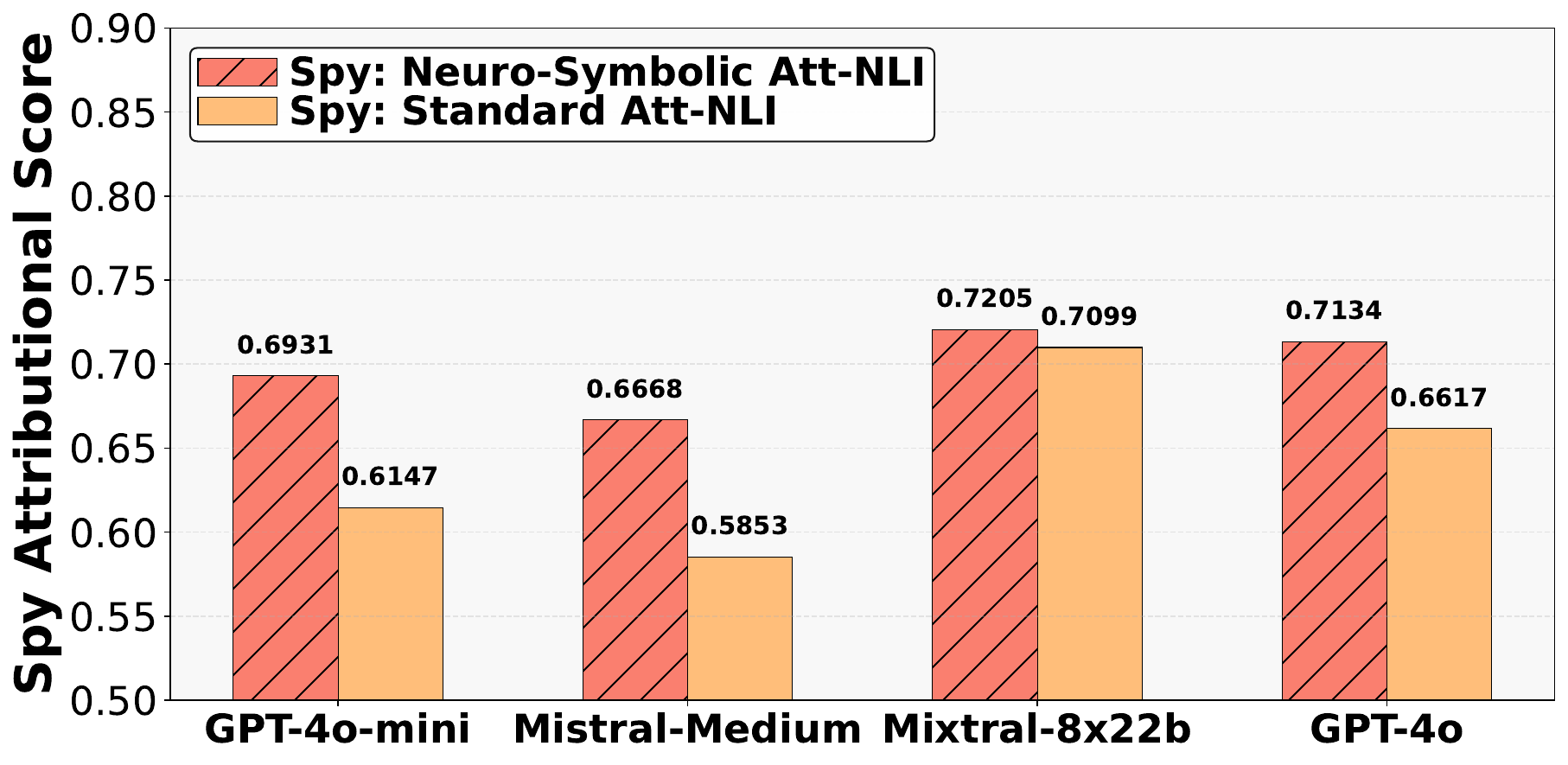}
        \caption{Spy attributional score}
        \label{fig:nmlsym_spy_score}
    \end{subfigure}
    \hfill
    \begin{subfigure}[b]{0.24\textwidth}
        \centering
        \includegraphics[width=\textwidth]{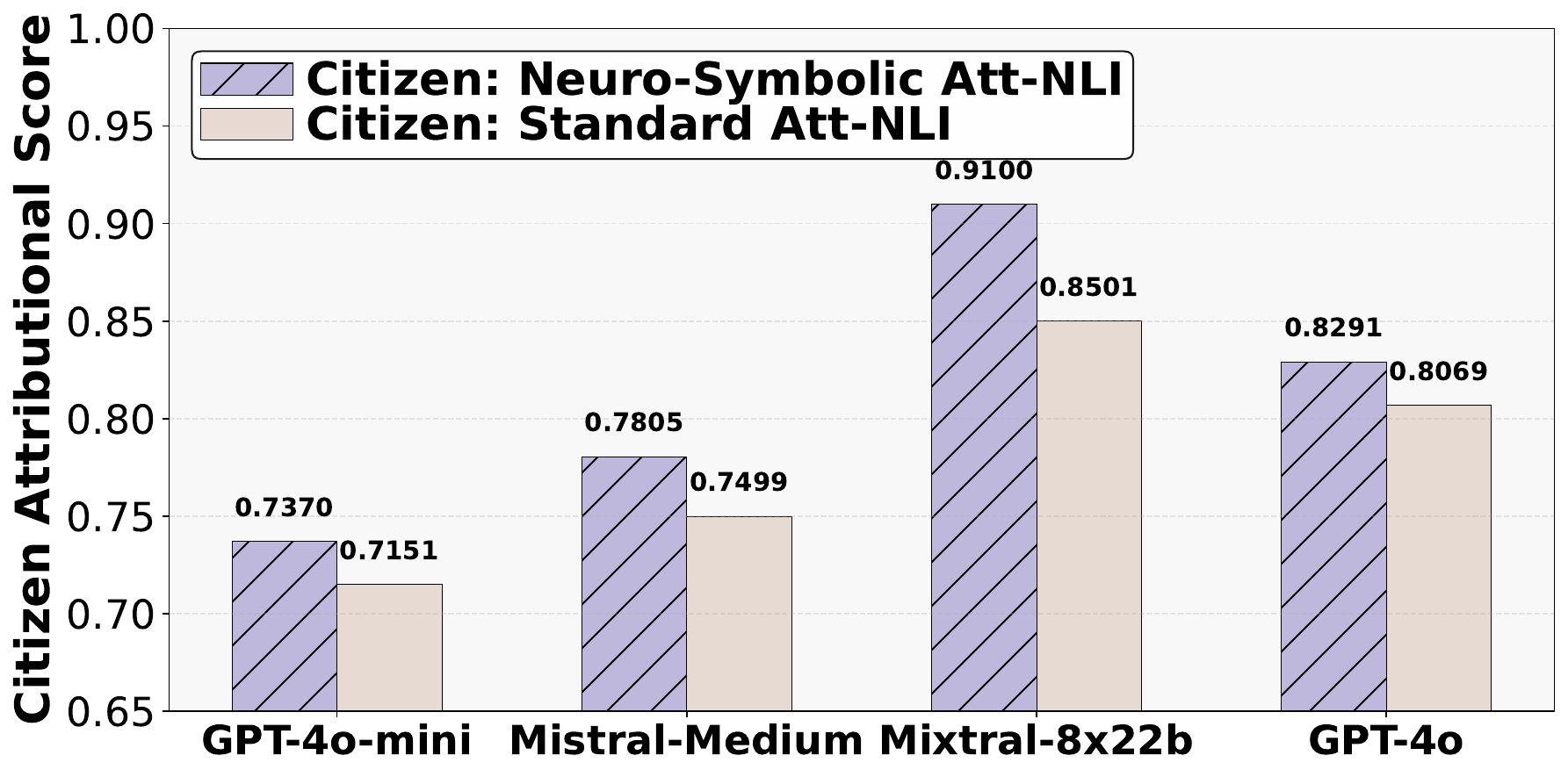}
        \caption{Cit. attributional score}
        \label{fig:nmlsym_citizen_score}
    \end{subfigure}
    \caption{Comparison between Standard Att-NLI and Neuro-Symbolic Att-NLI Player(1 Standard Att-NLI (Spy) vs. 5 Neuro-Symbolic (Cit.), 1 Neuro-Symbolic (Spy) vs. 5 Standard Att-NLI (Cit.)).}
    \label{fig:comparison_nml_sym}
\end{figure*}
\paragraph{LLMs exhibit various NLI and Att-NLI ability.} To investigate whether LLMs exhibit different NLI and Att-NLI abilities, we conduct experiments by setting GPT-4o-mini as the fixed citizen agent and assigning each LLM to the spy role as the target model. We run thirty independent games for each LLM's player type (standard NLI, standard Att-NLI, and neuro-symbolic Att-NLI) and perform a comprehensive analysis across different LLMs (Fig.~\ref{fig:comparison_baseline}). Fig.~\ref{fig:spy_win_baseline} shows that different LLMs exhibit varying win rates across all proposed player types when playing against the same citizen LLM. GPT-4o achieves the highest average win rates, while Mistral-Medium exhibits the lowest win rates in standard NLI, Att-NLI, and neuro-symbolic Att-NLI, respectively. Fig.~\ref{fig:spy_faithfullness_baseline} further illustrates that different LLMs have different attributional scores, with GPT-4o achieving the highest scores of 0.6, 0.688, and 0.780 in standard NLI, Att-NLI, and neuro-symbolic Att-NLI, respectively.

\paragraph{Att-NLI ability is critical in Undercover-V performance.} 
Fig.~\ref{fig:comparison_baseline} shows that a higher attributional score corresponds to stronger reasoning about other players' intentions and more valid conclusion inference, resulting in better game performance. The neuro-symbolic Att-NLI player achieves the highest attributional scores of 0.742, 0.664, 0.779, and 0.78, which align with the highest win rates of 20\%, 13.33\%, 20\%, and 33.33\% across all models, respectively. The consistency between attributional score and spy win rates demonstrates the critical importance of Att-NLI ability in the refined Undercover-V arena, which shows the effectiveness of Undercover-V as the arena for Att-NLI. Besides, we also researched the impact of word selection in Appendix~\ref{appendix:word selection}.

\subsection{Round-Robin Tournament}

Across the tournament, the neuro-symbolic Att-NLI player consistently outperforms the Standard Att-NLI player (Fig.~\ref{fig:comparison_nml_sym}) and surpasses the Standard NLI player (Fig.~\ref{fig:comparison_nve_sym}), while the Standard Att-NLI player in turn exceeds the Standard NLI player (Fig.~\ref{fig:comparison_nml_nve}). Accordingly, for every metric and role, performance and attributional score rank first for neuro-symbolic Att-NLI, second for Standard Att-NLI, and third for Standard NLI.

% \textbf{Att-NLI player can infer other players' intentions and generate more sound and valid inferences than the standard NLI player.}
% To investigate the differences between Att-NLI players and standard NLI players, we place them in a round-robin game, assigning each to the spy or the citizen role and running thirty games for each LLM and each mixed type. We evaluate standard Att-NLI and standard NLI using configurations of 1 Att-NLI vs. 5 NLI and 1 NLI vs. 5 Att-NLI, where the single player is the spy and the other five are citizens (see Fig.~\ref{fig:comparison_nml_nve}).
\paragraph{Att-NLI players infer others’ intentions and produce more sound and valid inferences than standard NLI players.} To examine these differences, we conduct round-robin games, assigning Att-NLI and standard NLI players to spy or citizen roles and running thirty games per LLM and mixed setting. We compare standard Att-NLI and standard NLI under 1 Att-NLI vs.\ 5 NLI and 1 NLI vs.\ 5 Att-NLI configurations, where the single player is the spy and the remaining five are citizens (see Fig.~\ref{fig:comparison_nml_nve}).

In games between standard Att-NLI and standard NLI, standard Att-NLI spies achieve higher win rates than standard NLI spies across all LLMs, with gains of 25.06\%, 66.70\%, 50.04\%, and 49.93\%. Their attributional scores are also higher, improving by 3.55\%, 0.34\%, 11.62\%, and 3.87\%. When playing as citizens, standard Att-NLI players are less likely to be eliminated, with elimination-rate reductions of 9.07\%, 17.64\%, 23.40\%, and 20.93\% compared to standard NLI citizens. They also attain higher overall attributional scores, with improvements of 3.38\%, 0.10\%, 4.05\%, and 2.73\% across different LLMs, respectively. In the round-robin tournament with neuro-symbolic players, the standard Att-NLI player achieves an average win rate of 13.75\%, compared to 9.58\% for standard NLI, indicating superior inference of other players’ intentions and identities and generation of descriptions closer to true citizen perspectives and more consistent with citizen descriptions.\\
\paragraph{Neuro-symbolic approach further enhances the Att-NLI ability.}
To investigate the differences between neuro-symbolic Att-NLI players, standard Att-NLI players, and standard NLI players, we assign each player type to either the spy or the citizen role and run thirty games for each LLM and each mixed type (see Fig.~\ref{fig:comparison_nml_sym} and \ref{fig:comparison_nve_sym}). In the game between neuro-symbolic Att-NLI and standard Att-NLI players, neuro-symbolic player achieves a higher win rate as spy than Att-NLI player's (see Figs.~\ref{fig:nmlsym_spy_win}, \ref{fig:nmlsym_citizen_eliminate}, \ref{fig:nmlsym_spy_score}, and \ref{fig:nmlsym_citizen_score}), with an average improvement of 32.08\% in win rate and 8.57\% in attributional score. As citizens, neuro-symbolic players have an average 7.22\% lower elimination rate and an average 3.97\% improvement in attributional score compared to standard Att-NLI players.\\
\paragraph{Feedback from external TP and refinement in the guess word significantly advances the Att-NLI ability.} By comparing the spy win rate in the overall round-robin tournament, we observe a significant average improvement (+78.29\%) for neuro-symbolic Att-NLI players over the standard NLI player, while standard Att-NLI players only show an enhancement of 43.53\% compared to the standard NLI player. A similar phenomenon appears in the citizen elimination rate, spy attributional score, and citizen attributional score (see Fig.~\ref{fig:nvenml_spywin},\ref{fig:nvenml_citizen_eliminate}, \ref{fig:nvenml_spy_score}, \ref{fig:nvenml_citizen_score},\ref{fig:nvesym_spywin}, \ref{fig:nvesym_citizen_eliminate},\ref{fig:nvesym_spy_score} and~\ref{fig:nvesym_citizen_score}). 
Among the evaluated LLMs, Mixtral-8x22b shows the most significant differences. Neuro-symbolic Att-NLI players demonstrate a 85.73\% improvement in spy win rate compared to standard NLI players, while standard Att-NLI players show a 57.18\% improvement. For the citizen elimination rate, neuro-symbolic Att-NLI players achieve a 27.99\% reduction compared to 15.99\% for standard Att-NLI. In the spy attributional score, neuro-symbolic Att-NLI players improve by 18.88\% versus 6.26\% for standard Att-NLI. Lastly, in the citizen attributional score, neuro-symbolic Att-NLI players improve by 10.18\% compared to 1.07\% for standard Att-NLI. These findings indicate that incorporating an external theorem prover enhances logical reliability during the inference stage. Moreover, feedback from the external theorem prover can be used to refine the guess word, which significantly helps the LLM agent infer and adjust to other agents' intentions, thus strengthening its Att-NLI performance.

\section{Related Work}

% \textbf{Natural Language Inference.}
% Numerous language models have been proposed and extensively evaluated on NLI tasks such as scientific question-answering~\citep{chen-etal-2017-enhanced,Liu2019RoBERTaAR,devlin-etal-2019-bert,NEURIPS2019_dc6a7e65}. More recently, research has advanced to multi-hop NLI and focused on generating explicit explanations that support or refute an entailment between a given premise and hypothesis~\citep{thayaparan-etal-2021-explainable, valentino2022hybrid, dalal-etal-2024-inference}. 
% These NLI methods are confined to single-agent settings and fail to infer latent intentions.

\paragraph{Neuro-symbolic Reasoning.}
Neuro-symbolic reasoning models integrate neural networks with symbolic solvers to provide a reliable and verifiable reasoning process for complex downstream tasks such as multi-hop reasoning~\citep{olausson-etal-2023-linc, pan-etal-2023-logic, jiang-etal-2024-leanreasoner, quan-etal-2024-verification, quan-etal-2025-peirce}. Research efforts have applied LLMs for autoformalization, converting natural language into first-order logic forms, and subsequently employing symbolic provers on logical reasoning tasks~\citep{olausson-etal-2023-linc, jiang-etal-2024-leanreasoner}. \citet{quan-etal-2024-verification, quan-etal-2025-faithful} integrated LLMs with external theorem provers for open-world NLI tasks to verify and refine natural language explanations. However, these methods are likewise confined to single-agent settings and cannot perform Att-NLI.\\
\paragraph{Multi-agent LLM Social Deduction Games.}
Recent work has explored the use of LLMs in multi-agent social deduction games as a benchmark for evaluating reasoning abilities~\citep{qiao2023gameevalevaluatingllmsconversational, costarelli2024gamebenchevaluatingstrategicreasoning}. These natural language conversation-based games often involve role play and deception. \citet{xu2024exploringlargelanguagemodels} proposed a framework using the Werewolf game to investigate how historical experiences affect LLMs' behaviors. \citet{10.5555/3692070.3694355} integrated LLMs with reinforcement learning policies to build agents for the Werewolf game. \citet{wang2023avalonsgamethoughtsbattle} introduced a recursive contemplation process, coupling an LLM with human-like recursive thinking and perspective-taking abilities in the Avalon game. However, lying and deception in these benchmarks make it conflict with the principle of explosion (\textit{ex falso [sequitur] quodlibet}) for NLI~\citep{Smith2003-SMIAIT-2}.

\section{Conclusion}
% We propose Attributional NLI (Att-NLI) to extend standard NLI to multi-agent LLMs' intention inference and introduce an empirically testable game (Undercover-V) with a novel evaluation metric to assess this ability. We conduct comprehensive experiments on word selection, fixed-opponent contests, and round-robin tournaments, showing that our three proposed agent types demonstrate progressively enhanced Att-NLI. Among these, the neuro-symbolic agents achieve the best overall reasoning performance in spy win rate and the citizen elimination rate, evidenced by the highest attributional scores. Our framework and experimental findings will provide a strong foundation for future research on Att-NLI in LLM-based multi-agent studies.

We propose Attributional NLI (Att-NLI), extending standard NLI to intention inference in multi-agent LLMs, and introduce an empirically testable game (Undercover-V) with a novel metric. Extensive experiments on word selection, fixed-opponent contests, and round-robin tournaments show progressively stronger Att-NLI across three agent types. Neuro-symbolic agents achieve the best overall reasoning performance in spy win rate and citizen elimination rate, as reflected by the highest Attributional Scores. Our framework and findings provide a strong foundation for future study on Att-NLI in LLM-based multi-agent settings.

\section*{Limitations}
Our proposed framework introduces attributional natural language inference (Att-NLI) through the social-deduction game Undercover-V as a minimum viable testbed to evaluate intentional reasoning in LLM agents, but it also relies on a number of structural simplifications that constrain the scope of our conclusions. At a conceptual level, Undercover-V reduces attribution to discriminating between two latent roles (spy vs.\ citizen) that are tightly tied to a pair of lexical anchors, so that “intention” is proxied by which of two nearby word-level concepts best explains an agent’s utterances; even though we carefully control word-pair difficulty and analyse embedding-based similarity and bias, this design still privileges local lexical semantics over richer, temporally extended behaviour or social norms, which are central to attribution in real multi-agent systems. Our overall framework presupposes that attributional competence can be meaningfully assessed via performance in a cooperative–adversarial guessing game and an embedding-based Attributional Score; this operationalisation is useful for isolating a specific facet of intention-aware reasoning, but it does not yet address how Att-NLI interacts with broader desiderata such as social calibration, trustworthiness, safety, or alignment with human judgments in more open-ended multi-agent environments.

%\section*{Acknowledgments}

%This work was partially funded by the SNSF project NeuMath (\href{https://data.snf.ch/grants/grant/204617}{200021\_204617}), by the EPSRC grant EP/T026995/1, “EnnCore” under Security for all in an AI-enabled society, by the CRUK National Biomarker Centre, and supported by the Manchester Experimental Cancer Medicine Centre and the NIHR Manchester Biomedical Research Centre.


\bibliography{custom}

\appendix

\section{Undercover-V Game Rules}
\label{appednix:game_rule}
Undercover-V is a social deduction game involving six players, one of whom 
is designated as the spy. Each player privately receives a ``word card'': five players hold the same word (e.g., ``banana''), while the spy holds 
a different one (e.g., ``apple''). During each description phase, 
every participant gives a single-sentence description of their word one by one. 
These descriptions must not contradict the assigned card, nor repeat 
any previously stated clues. Only the resulting sentences are revealed 
publicly.

Following the description phase, players enter a voting phase in which all six participants cast a vote for the person they suspect to be the spy at the same time. Abstentions are disallowed, and the individual receiving the most votes is immediately eliminated. If a player holds the spy word voted out at any point, the citizen players collectively win. If the player who holds the spy word survives to the last round (only two players alive), the spy player immediately wins. In the event of a tied vote for three consecutive rounds, the spy is declared the winner, as the group fails to converge on a suspect.

The word pairs we selected are: Lip balm-Lip cream, Imagination-Imaginary, Cherry Blossom-Peach Blossom, Ophiophagus-Naja, Earl Grey Tea-Ceylon Tea, Sweet Orange-Navel Orange, Ethics-Morality, Impatiens hawkeri-Impatiens walleriana, Filistatidae-Hypochilidae, and Saussurella-Tettigidea.

\section{Implementation of External Theorem Prover in Neuro-Symbolic Player}
\label{appendix:isabelle}
We integrate Isabelle/HOL~\citep{nipkow2002isabelle} as an external theorem prover, in conjunction with LLMs, to perform verification and refinement of the autoformalized Isabelle/HOL theory. The details are as follows: 

\subsection{Mechanism of integrating external theorem prover for verification and refinement} 

\subsubsection{Verification}
(i) Knowledge Base Construction. Let $D$ be the set of all players' historical descriptions. For each player $i$, 
we define \(D_i = \{d_i^{(r)} | r \in \mathbb{N}^+\} \subseteq D\), where $d_i^{(r)}$ is the description generated by player $i$ in round $r$. We treat \(D_i\) as a set of facts \(F_i\) about player $i$'s past statements. An LLM is then prompted with \(p_{\text{NAtt}}^{(\star)}\) (same for both description and voting phases) to produce a set of rules \(R_i\) based on these facts: \(R_i = \mathrm{LLM}\bigl(F_i, p_{\text{NAtt}}^{(\star)}\bigr).\)
Subsequently, we construct the knowledge base \(\mathcal{KB}_i\) by combining the original facts \(F_i\) with the newly generated rules \(\mathcal{KB}_i = F_i \cup R_i\), yielding a set of factual statements and inferred rules.

(ii) Autoformalization and Verification.
Next, the natural language sentences in \(\mathcal{KB}_i\) are autoformalized into logical forms (Neo-Davidsonian event semantics \citep{Parsons1990EventsIT}) within Isabelle/HOL, yielding a set of axioms \(A = \{a_1, a_2,\ldots, a_m\}\). Meanwhile, the neuro-symbolic player's own word card \(w_i\) is formalized into a theorem \(\tau\). Combining the axioms and the theorem yields a theory \(\Theta = (A, \tau)\), which is passed to Isabelle/HOL for a \textit{early-stop majority vote} automated verification. If a syntax error is identified from the TP, we label \(d_i\) as a syntax error. Otherwise, if no proof is found, we label \(d_i\) logically invalid; if a proof is found, we deem it logically valid. We repeat this process for each player \(i\) in the game, ultimately collecting all verification outcomes in a record \(\mathcal{V}\), where each entry is classified as logically valid, logically invalid, or syntax error.

\subsubsection{Refinement}
While the logical record \(\mathcal{V}\) augments the standard Att-NLI player's abductive-deductive inference, we further propose a theory-driven correction for neuro-symbolic players. After each voting phase, a neuro-symbolic player refines its previously guessed opponent's word \(g_i\). Let the voted-out player in round \(r\) be \(j\). If \(d_j^{(r)}\) (the voted-out player's description) 
is labeled logically invalid in \(\mathcal{V}\) (indicating it does not logically entail the player's card \(w_i\) that the voted-out player might be an opponent player), we construct a new theory \(\Theta' = (A', \tau')\), where \(A'\) is derived from the voted-out player's descriptions, and \(\tau'\) is the target theorem derived from the guessed 
word \(g_i\). If the theorem prover (TP) indicates no valid proof for \(\Theta'\), we obtain a mismatch between \(d_j^{(r)}\) and \(g_i^{(r)}\), suggesting that the guess may be wrong. 
Formally, define an indicator function

\begin{equation}
    \mathrm{valid}(\Theta')
    =
    \begin{cases}
      1, & \text{if $\Theta'$ is provable by TP},\\[2pt]
      0, & \text{otherwise}.
    \end{cases}
\end{equation}
We then apply the following piecewise rule to update (or retain) the guess \(g_i\). Let \(\theta_i^{(r)} \in \{\text{spy},\text{citizen}\}\) be the neuro-symbolic player's identity hypothesis in round \(r\) and \(\mathrm{Traces}(\Theta')\) denoted by the trace information (erroneous proof steps) extracted from \(\Theta'\) if no proof is found. 
The new guess word \(g_i^{(r+1)}\) is determined by
\begin{equation}
g_i^{(r+1)}=
\begin{cases}
\begin{aligned}[t]
\mathrm{LLM}\bigl(
   g_i^{(r)},\,\\
   \mathrm{Traces}(\Theta'),\\
   \theta_i^{(r)},\,
   p_{\text{up}}
\bigr)
\end{aligned}
& \text{if }\mathrm{valid}(\Theta')=0,\\[6pt]
g_i^{(r)} & \text{otherwise}.
\end{cases}
\end{equation}

\begin{figure*}[!t]
\centering
\begin{equation}
    g_i^{(r+1)} 
    =
    \begin{cases}
        \mathrm{LLM}\bigl(g_i^{(r)},\mathrm{Traces}(\Theta'),\theta_i^{(r)},p_{\text{up}}\bigr), 
          & \text{if } \theta_i^{(r)}=\text{spy} \,\wedge\, \mathrm{valid}(\Theta')=0,\\[6pt]
        g_i^{(r)}, 
          & \text{if } \theta_i^{(r)}=\text{citizen} \,\wedge\, \text{player opts to keep the guess},\\[6pt]
        \mathrm{LLM}\bigl(g_i^{(r)},\mathrm{Traces}(\Theta'),\theta_i^{(r)},p_{\text{up}}\bigr),
          & \text{if } \theta_i^{(r)}=\text{citizen} \,\wedge\, \text{the logical error is severe}.
    \end{cases}
\label{update_equation}
\end{equation}
\end{figure*}
In words, if \(\Theta'\) is unprovable, the player extracts proof errors via \(\mathrm{Traces}(\Theta')\) and uses them as feedback to the LLM (with update guess word prompt \(p_{\text{up}}\)) to refine its guess. Conversely, if \(\Theta'\) is 
provable, the guess is retained. 
Furthermore, the player's identity \(\theta_i^{(r)}\) can influence the decision to trigger an update as shown in Eq. \ref{update_equation}.

Here, the bottom two branches illustrate that a citizen might \emph{keep} the guess (if they believe the majority incorrectly voted out a valid player) or \emph{update} the guess if the logical errors are too significant to ignore.

\subsection{Constructing the Knowledge Base}
Since the knowledge base \(\mathcal{KB}_i\) is constructed by combining the factual statements (player descriptions \(D_i\)) with newly generated rules from the LLM that \(\mathcal{KB}_i = F_i \cup R_i\). Specifically, the rules generated by the LLM align with real-world knowledge and provide generalized statements based on the facts \(F_i\). For example, given the fact: ``An insect is known for its chirping sound, often heard in the evening'', the LLM might generate rules such as: ``Saussurella is a type of cricket known for producing a chirping sound.'' and ``If an insect is known for its chirping sound, often heard in the evening, it might be a Saussurella.''

\subsection{Autoformalization}
Following \citet{quan-etal-2024-verification}, we adopt Neo-Davidsonian event semantics \citep{Parsons1990EventsIT} as the logical form to represent the natural language sentences in \(\mathcal{KB}_i\). This approach encodes semantic roles in predicate-argument structures, capturing the event verbs and relevant participants (agents, patients, etc.) more faithfully. As a result, it helps preserve semantic information during autoformalization and remains closer to the surface form of the original sentences \citep{quan-etal-2024-enhancing, quan-etal-2024-verification}.

For example, consider the sentence: wolf is an animal that does not live in the trees. 
We formalize it as:  
\begin{equation}
\begin{aligned}
  \forall x y e.\Bigl(
    \text{Animal}(x)
    &\wedge \text{Tree}(y)
    \wedge \neg\text{LiveOn}(e)\\
    &\wedge \text{Agent}(e,x)
    \wedge \text{Patient}(e,y)
  \Bigr)\\
    &\to \exists z.\bigl(\text{Wolf}(z)\wedge z = x\bigr).
\end{aligned}
\end{equation}
In this representation, \(\text{Animal}(x)\) serves as the agent of the 
non-\emph{LiveOn} event, and \(\text{Tree}(y)\) is the patient. By specifying 
\(\neg\text{LiveOn}(e)\), the formalization indicates that the agent \(x\) 
(an animal) does not live on the patient \(y\) (the tree). The existential 
clause \(\exists z.\text{Wolf}(z)\wedge z = x\) then asserts that \(x\) 
must be a wolf. Hence, the formalization faithfully captures the original 
sentence's semantics while retaining explicit event structures.

Specifically, the natural language sentences in \(\mathcal{KB}_i\) are converted
into Neo-Davidsonian event semantics using an LLM. Let \(\Phi\) be the function 
that performs this transformation:
\begin{equation}
  \Phi\bigl(nl\bigr) 
  = 
  \mathrm{LLM}\bigl(nl, p_{\text{davidsonian}}\bigr),
  \quad nl \in \mathcal{KB}_i,
\end{equation}
where \(p_{\text{davidsonian}}\) denotes the prompt used for logical-form 
transformation. We then define \(\phi\) as the set of all resulting logical forms:
\begin{equation}
  \phi 
  = 
  \{\Phi\bigl(nl\bigr)| nl \in \mathcal{KB}_i\}.
\end{equation}
After obtaining the autoformalized logical forms set $\phi$, we proceed to construct the Isabelle theory $\Theta$. This theory consists of a set of axioms $A$ and a theorem $\tau$. We formulate the axioms $A$ of $\Theta$ as follows:
$$A = {a_1, a_2, ..., a_n}$$
where each axiom $a_i$ corresponds to a fact or a rule in logical forms \(\phi\), and is derived using an LLM:
$$a_i = \text{LLM}(p_{\text{axiom}}, \phi)$$
Here, $p_{\text{axiom}}$ denotes the prompt used for transforming logical forms into Isabelle/HOL axioms code.
The theorem \(\tau\) is then constructed according to the player's holding word 
card \(w_i\). Fig.~\ref{isabelle_example} shows an example of a constructed Isabelle/HOL theory.

\begin{figure*}
\begin{lstlisting}
begin

typedecl entity
typedecl event

consts
  Tea :: "entity $\Rightarrow$ bool"
  Bergamot :: "entity $\Rightarrow$ bool"
  Aroma :: "entity $\Rightarrow$ bool"
  Infuse :: "event $\Rightarrow$ bool"
  Agent :: "event $\Rightarrow$ entity $\Rightarrow$ bool"
  Patient :: "event $\Rightarrow$ entity $\Rightarrow$ bool"
  Give :: "event $\Rightarrow$ bool"
  Distinct :: "entity $\Rightarrow$ bool"
  Citrusy :: "entity $\Rightarrow$ bool"
  CeylonTea :: "entity $\Rightarrow$ bool"
  BoldFlavor :: "entity $\Rightarrow$ bool"
  Bright :: "entity $\Rightarrow$ bool"
  KnowFor :: "event $\Rightarrow$ bool"
  Typically :: "event $\Rightarrow$ bool"
  EarlGrey :: "entity $\Rightarrow$ bool"

(* Fact 1: Often infused with bergamot, giving it a distinct citrusy aroma *)
axiomatization where
  fact_1: "$\forall$x y z e1 e2. Tea x $\wedge$ Bergamot y $\wedge$ Aroma z $\wedge$ Infuse e1 $\wedge$ Agent e1 x $\wedge$ Patient e1 y $\wedge$ Give e2 $\wedge$ Agent e2 x $\wedge$ Patient e2 z $\wedge$ Distinct z $\wedge$ Citrusy z"

(* Rule 1: Ceylon Tea is known for its bold flavor and bright, citrusy aroma, but it is not typically infused with bergamot *)
axiomatization where
  rule_1: "$\forall$x y z e1 e2. CeylonTea x $\wedge$ BoldFlavor y $\wedge$ Aroma z $\wedge$ Bright z $\wedge$ Citrusy z $\wedge$ KnowFor e1 $\wedge$ Agent e1 x $\wedge$ Patient e1 y $\wedge$ Patient e1 z $\wedge$ Bergamot w $\wedge$ Infuse e2 $\wedge$ Agent e2 x $\wedge$ Patient e2 w $\rightarrow$ $\neg$Typically e2"

(* Rule 2: If a tea is often infused with bergamot, it is more likely to be Earl Grey rather than Ceylon Tea *)
axiomatization where
  rule_2: "$\forall$x y e. Tea x $\wedge$ Bergamot y $\wedge$ Infuse e $\wedge$ Agent e x $\wedge$ Patient e y $\wedge$ Often e $\rightarrow$ ($\exists$z. EarlGrey z $\wedge$ z = x) $\wedge$ $\neg$($\exists$w. CeylonTea w $\wedge$ w = x)"


theorem hypothesis:
  shows "$\exists$x. CeylonTea x"
  sledgehammer
  oops

end
\end{lstlisting}
\caption{An example of the constructed Isabelle/HOL Theory.} 
\label{isabelle_example}
\end{figure*}

\subsection{Verification}
As autoformalization may introduce syntactic errors when transferring natural language 
into Isabelle/HOL code, we employ the iterative approach of \citet{quan-etal-2024-verification, quan-etal-2025-faithful} 
to detect and refine such errors. Let \(E(\Theta)\) be a function that detects syntax 
errors using a theorem prover:
\begin{equation}
  E(\Theta) = TP(\Theta),
\end{equation}
where \(TP\) represents the theorem prover. If \(E(\Theta)\neq\emptyset\), indicating 
the presence of error messages, we use an LLM to refine these errors:
\begin{equation}
  \Theta' = \mathrm{LLM}\bigl(p_{\text{syntax}}, \Theta, E(\Theta)\bigr),
\end{equation}
where \(p_{\text{syntax}}\) is the prompt for syntax error refinement, and \(\Theta'\) 
is the refined theory. Algorithem \ref{algorithm_1} shows the detailed algorithm of how to implement the syntax error refinement process.

\begin{algorithm}
    
    \SetKwFunction{isOddNumber}{isOddNumber}
    
    \SetKwInOut{KwIn}{Input}
    \SetKwInOut{KwOut}{Output}

    \KwIn{Isabelle/HOL theory $\Theta$, Isabelle//HOL server $isabelle$, Syntax refinement model $m_s$}
    \KwOut{Isabelle/HOL theory $\Theta$}
    $iterations \leftarrow 0$  \\
    $max\_iterations \leftarrow 5$ \\
    $has\_syntax\_error$ $\leftarrow$ $true$ \\
    \While{ $has\_syntax\_error$ {\bf and} $iterations < max\_iterations$}{
    session\_id $\leftarrow$ session\_build($HOL,isabelle$) \\
        $isabelle$.start(session\_id) \\
    $isabelle\_response$ $\leftarrow$ $isabelle$.check($\Theta$)\\
    \eIf{$syntax\_errors$ in $isabelle\_response$}{
     $\Theta$ $\leftarrow$ refine\_syntax($syntax\_errors$, $\Theta$,$m_s$)\\
     $it$ $\leftarrow$ $it+1$
    }{
    $has\_syntax\_error$ $\leftarrow$ $false$
    }
    }
    \KwRet{$\Theta$}
    \caption{Syntax Error Refinement}
    \label{algorithm_1}
\end{algorithm}

\begin{figure*}
\begin{lstlisting}
theorem hypothesis:
  shows "$\exists$x. GreenTea x"
  proof -
  have "$\exists$y. Popular y $\wedge$ BlackTea y $\wedge$ (SliceOfLemon y $\vee$ ($\exists$z. SplashOfMilk z $\wedge$ EnjoyedWith y z))" sledgehammer
  then have "$\neg$($\exists$x. GreenTea x)" <ATP>
  then show ?thesis <ATP>
qed
  
\end{lstlisting}
\caption{An example of the proof sketch with <ATP> tags.} 
\label{proof_sketch_example}
\end{figure*}

After the syntax check and refinement stage, we use the Sledgehammer tool \citep{Paulson2012ThreeYO} in Isabelle to invoke external automated theorem provers (ATPs) in search of a valid proof. Formally, we let
\begin{equation}
  V_{\text{sledge}}(\Theta) = \mathrm{Sledgehammer}(\tau),
\end{equation}
where \(V_{\text{sledge}}(\Theta)\) denotes the ATP search result of Sledgehammer's attempt on the constructed theory \(\Theta = (A, \tau)\). 

If Sledgehammer successfully finds a proof, we extract all possible proofs from its results and conclude that the explanation is logically valid. Otherwise, we prompt an LLM to generate a proof sketch, which is an outline describing how the theorem might be proved. This sketch is then submitted to the theorem prover for step-by-step verification. If the prover fails at any step, we identify that specific step as erroneous and return it as corrective feedback. Fig.~\ref{proof_sketch_example} illustrates an example of a generated proof sketch in which \texttt{<ATP>} tags are iteratively replaced by calls to Sledgehammer until the first failure occurs.
\begin{algorithm}[t!]
\SetKwInOut{KwIn}{Input}
\SetKwInOut{KwOut}{Output}

\KwIn{Isabelle/HOL theory $\Theta$, Isabelle/HOL server $isabelle$}
\KwOut{Final outcome (\emph{valid}, \emph{invalid}, or \emph{syntax error})}

Initialize counters: $cnt[\text{valid}] = 0, cnt[\text{invalid}] = 0, cnt[\text{syntax\_error}] = 0$\\
\While{true}{
    $r \leftarrow isabelle.\textit{check}(\Theta)$\\
    $cnt[r] \leftarrow cnt[r] + 1$\\
    \If{$cnt[r] = 2$}{
        \KwRet{$r$}
    }
}
\caption{Early-Stop Majority Vote}
\label{algorithm_2}
\end{algorithm}

\textbf{Early-Stop Majority Vote.}
Since the theorem prover may produce inconsistent results (e.g., valid, invalid, or syntax error) across multiple runs of the same theory, 
we adopt an early-stop majority vote procedure. Specifically, we repeat 
the proof attempt several times; each attempt yields one of three outcomes 
(valid, invalid, or syntax error). We maintain a count 
for each outcome. Once any outcome appears twice, we immediately choose 
it as the final logical result for the theory without further iterations. This 
strategy allows us to settle on a result quickly in cases where the theorem 
prover exhibits occasional randomness or system-level inconsistencies.

\subsection{Scalability of the theorem prover}
\label{scalability_isabelle}
\begin{figure}[t]
    \centering
    \includegraphics[width=0.5\textwidth]{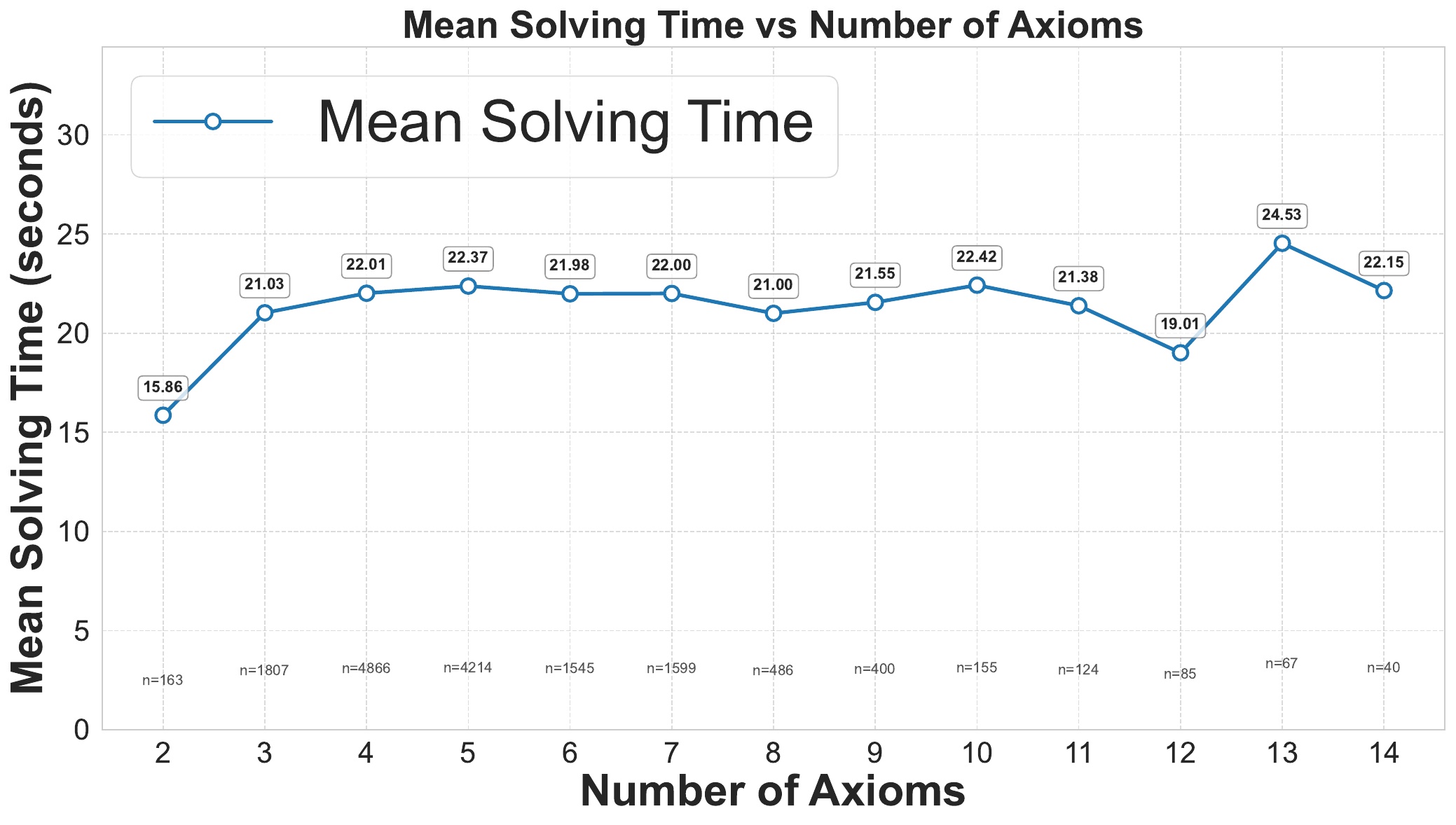}
    \caption{The average solving time against number of axioms in the Isabelle/HOL theory} 
    \label{fig:scalability}
\end{figure}

Fig. \ref{fig:scalability} shows the average solving time versus the number of axioms in the Isabelle/HOL theory for integrating Isabelle/HOL with LLMs in neuro-symbolic players.

\section{Formalization and Proof of Testability in Undercover-V}
\label{appendix:proof}
Consider a game with players 
\begin{equation}
\mathcal{P} = \{1, 2, \dots, n\},
\end{equation}
where each player \(i\) is assigned a hidden state \(s_i \in \mathcal{S}\) with 
\begin{equation}
\mathcal{S} = \{s_c, s_s\},
\end{equation}
and the constraint that exactly one player holds the state \(s_s\):
\begin{equation}
\left|\{ i | s_i = s_s \}\right| = 1.
\end{equation}

For each \(s \in \mathcal{S}\), let \(\phi(s)\) be a fixed set of axioms in a first-order language \(\mathcal{L}\) that encodes the \emph{verifiable features} of \(s\). (The precise construction of \(\phi(s)\) is domain-specific; it is assumed to be given by expert knowledge so that \(\phi(s)\) represents the necessary facts that must hold for any player with state \(s\).)

In each round \(r\), every player \(i\) produces a natural language description \(d_i^{(r)}\). Assume that there exists a parsing function
\begin{equation}
\text{parse}: d_i^{(r)} \mapsto \Delta_i^r \in \mathrm{Form}(\mathcal{L}),
\end{equation}
which maps the natural language description \(d_i^{(r)}\) to a well-formed formula (or set of formulas) \(\Delta_i^r\) in \(\mathcal{L}\). (Here, \(\mathrm{Form}(\mathcal{L})\) denotes the set of all well-formed formulas in \(\mathcal{L}\).)

We define the test function 
\begin{equation}
T(s,\Delta) = \text{True} \quad \Longleftrightarrow \quad \phi(s) \cup \{\Delta\} \not\models \bot,
\end{equation}
where \(\bot\) denotes a contradiction. That is, \(T(s,\Delta)=\text{True}\) if and only if the formula \(\Delta\) is logically consistent with the axioms \(\phi(s)\) of state \(s\). In our game (Undercover-V), the rules require that for every player \(i\) and every round \(r\),
\begin{equation}
T(s_i,\Delta_i^r)=\text{True}.
\end{equation}
This means that the parsed description \(\Delta_i^r\) must be consistent with \(\phi(s_i)\); both partial inconsistencies (where only some components conflict with \(\phi(s_i)\)) and complete inconsistencies are forbidden. Note that this condition does not require a player to fully specify all of \(\phi(s_i)\); ambiguous or vague descriptions are acceptable as long as they do not lead to a contradiction.

We now show that, under these rules, for any player \(i\) and any number of rounds \(R\), the cumulative set of descriptions 
\begin{equation}
\Delta_i^{(R)} = \bigcup_{r=1}^{R}\Delta_i^r
\end{equation}
remains consistent with \(\phi(s_i)\), i.e.,
\begin{equation}
\phi(s_i)\cup\Delta_i^{(R)} \not\models \bot.
\end{equation}
This cumulative consistency is what we refer to as the testability of the game, and it guarantees that an external verifier can, by applying logical inference to the accumulated information, eventually converge on the true state \(s_i\) of each player.

\textbf{Proof (by induction on \(R\)):}

\textbf{Base Case (\(R=1\)):}  
By the game rule, for every player \(i\),
\begin{equation}
T(s_i,\Delta_i^1)=\text{True}.
\end{equation}
Hence,
\begin{equation}
\phi(s_i)\cup\{\Delta_i^1\} \not\models \bot.
\end{equation}

\textbf{Inductive Step:}  
Assume that after \(R\) rounds,
\begin{equation}
\phi(s_i)\cup \bigcup_{r=1}^{R}\{\Delta_i^r\} \not\models \bot.
\end{equation}
In round \(R+1\), the game rule enforces that
\begin{equation}
T(s_i,\Delta_i^{R+1})=\text{True},
\end{equation}
so that
\begin{equation}
\phi(s_i)\cup\{\Delta_i^{R+1}\} \not\models \bot.
\end{equation}
Assuming that the parsing function and the construction of \(\phi(s_i)\) do not introduce extraneous contradictions, the union of the consistent sets
\(\phi(s_i)\cup \bigcup_{r=1}^{R}\{\Delta_i^r\}\) and \(\{\Delta_i^{R+1}\}\) remains consistent. Hence,
\begin{equation}
\phi(s_i)\cup\left(\bigcup_{r=1}^{R}\{\Delta_i^r\} \cup \{\Delta_i^{R+1}\}\right) \not\models \bot.
\end{equation}
Thus, by induction, for all \(R\ge 1\),
\begin{equation}
\phi(s_i)\cup\Delta_i^{(R)} \not\models \bot.
\end{equation}

This cumulative consistency enables an external verifier to apply logical inference methods over the accumulated descriptions to gradually eliminate inconsistent hypotheses about \(s_i\) and ultimately converge on the actual state.

By contrast, in a game that allows unrestricted lying (e.g., Werewolf), there exists at least one player \(p_j\) and round \(r\) such that
\begin{equation}
\phi(s_j)\cup\{\Delta_j^r\} \models \bot,
\end{equation}
meaning that a player's statement contradicts the verifiable features of their true state. Accumulating such contradictory statements prevents reliable logical convergence on the true state.

Thus, Undercover-V is testable in that every player's statement is required to be consistent with their state-specific axioms, ensuring that, over successive rounds, the aggregated information remains logically coherent and permits correct identification of each player's true state.

\begin{table*}[t]
\centering
\scalebox{0.6}{
\begin{tabular}{p{4cm} p{20cm}}
\toprule
\rowcolor{gray!20} 
\multicolumn{2}{c}{\textbf{Description Phase (One after another)}}\\
\midrule
Player 1 (Earl Grey Tea)& \textbf{Intention Selection:} Since this is the first round and there are no other players' descriptions, I cannot deduce if I am the spy or not.\newline \textbf{Conclusion Inference:} I will describe my word card as accurately as possible, while being careful not to reveal too much information. \newline \textbf{Description:}A type of tea known for its distinct flavor and named after a British nobleman.\\
\midrule
Player 2 (Earl Grey Tea)& \textbf{Intention Selection:} Player 1's description could potentially deduce my word card, as it describes a type of tea. However, it does not specifically mention Earl Grey Tea.\newline \textbf{Conclusion Inference:} Since I am not certain if I am the spy or not, I will describe my word card in a way that aligns with Player 1's description, but does not explicitly reveal my word. \newline \textbf{Description:}A type of tea that is often enjoyed with a slice of lemon and named after a British nobleman. \\
\midrule
Player 3 (Earl Grey Tea)& \textbf{Intention Selection:} Both Player 1 and Player 2 have described their word as a type of tea named after a British nobleman. This aligns with my word card, Earl Grey Tea. Therefore, I am not the spy.\newline \textbf{Conclusion Inference:} Since I am not the spy, I will describe my word card in a way that aligns with the other players' descriptions, while ensuring it is unique. \newline \textbf{Description:}A type of tea that is often served with milk and sugar and named after a British nobleman. \\
\midrule
Player 4 (Earl Grey Tea)& \textbf{Intention Selection:} All descriptions provided by the other players seem to be describing a type of tea named after a British nobleman. This aligns with my word card, Earl Grey Tea. Therefore, I am likely not the spy.\newline \textbf{Conclusion Inference:} Since I am not the spy, I will describe my word card in a way that aligns with the other players' descriptions, while ensuring my description is unique. \newline \textbf{Description:}A type of tea that is often enjoyed in the afternoon and named after a British nobleman.  \\
\midrule
Player 5 (Earl Grey Tea)& \textbf{Intention Selection:} All descriptions provided by the other players seem to be describing a type of tea named after a British nobleman. This aligns with my word card, Earl Grey Tea. Therefore, I am likely not the spy.\newline \textbf{Conclusion Inference:} Since I am not the spy, I will describe my word card accurately and in a way that fits with the other players' descriptions. \newline \textbf{Description:}A type of tea that is often enjoyed with a splash of milk and named after a British nobleman.  \\
\midrule
\rowcolor{red!20} 
Player 6 (Ceylon Tea)& A type of tea that originates from the island formerly known as Ceylon, now Sri Lanka. \\
\midrule
\rowcolor{gray!20} 
\multicolumn{2}{c}{\textbf{Voting Phase (At the same time)}}\\
\midrule
Player 1 (Earl Grey Tea) & \textbf{Intention Selection:} All the other players' descriptions seem to align with my word card, I am not the spy. \newline \textbf{Conclusion Inference:} Player 6's description does not mention a British nobleman, which is a key characteristic of Earl Grey Tea. \newline \textbf{Vote:}I will vote for player 6.\\
\midrule
Player 2 (Earl Grey Tea)& \textbf{Intention Selection:} The majority of the descriptions align with my word card, Earl Grey Tea. However, Player 6's description does not match the others, as it mentions the origin of the tea rather than its association with a British nobleman. \newline \textbf{Conclusion Inference:} Since I am not the spy, I will vote for Player 6 as the spy, as their description does not align with the majority of the descriptions. \newline \textbf{Vote:}I will vote for player 6. \\
\midrule
Player 3 (Earl Grey Tea)& \textbf{Intention Selection:} All the other players' descriptions align with my word card, Earl Grey Tea, I am not the spy.  \newline \textbf{Conclusion Inference:} Player 6's description mentions the origin of the tea, which does not align with my word card. Therefore, Player 6 is likely the spy. \newline \textbf{Vote:}I will vote for player 6.\\
\midrule
Player 4 (Earl Grey Tea)& \textbf{Intention Selection:} All descriptions provided by the other players seem to be describing a type of tea named after a British nobleman, except for Player 6's description. This aligns with my word card, Earl Grey Tea. Therefore, I am likely not the spy, and Player 6 is the most suspicious. \newline \textbf{Conclusion Inference:} I will vote for Player 6 as the spy, as their description does not align with the majority of the other players' descriptions. \newline \textbf{Vote:}I will vote for player 6.\\
\midrule
Player 5 (Earl Grey Tea)& \textbf{Intention Selection:} All descriptions provided by the other players seem to be describing a type of tea. However, Player 6's description does not align with the others, as it does not mention a British nobleman. Therefore, Player 6 is likely the spy. \newline \textbf{Conclusion Inference:} Since I am not the spy, I will vote for Player 6 as the spy. \newline \textbf{Vote:}I will vote for player 6. \\
\midrule
\rowcolor{red!20} 
Player 6 (Ceylon Tea)& I will vote for Player 5. \\
\bottomrule
\end{tabular}
}%
\caption{A round-robin tournament game ended after a single round in which a Standard NLI (Player 6) played against five Standard Att-NLI players. During the description phase, each player gave their description in turn. After the description phase, all players cast their votes simultaneously.}
\label{case_study}
\end{table*}

\begin{figure*}[htbp]
    \centering
    \begin{subfigure}[b]{0.24\textwidth}
        \centering
        \includegraphics[width=\textwidth]{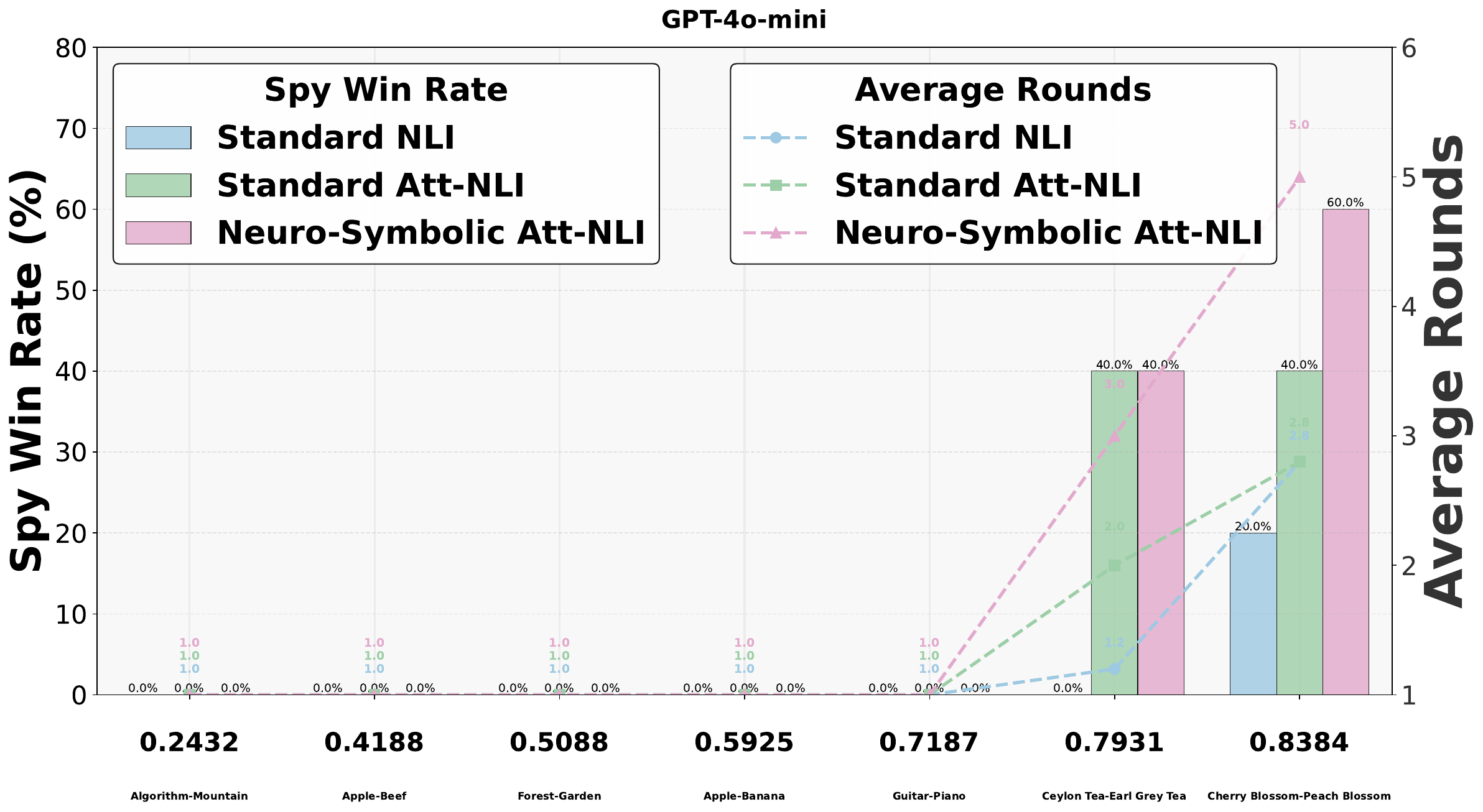}
        \caption{GPT-4o-mini}
        \label{fig:4omini_word_sim}
    \end{subfigure}
    \hfill
    \begin{subfigure}[b]{0.24\textwidth}
        \centering
        \includegraphics[width=\textwidth]{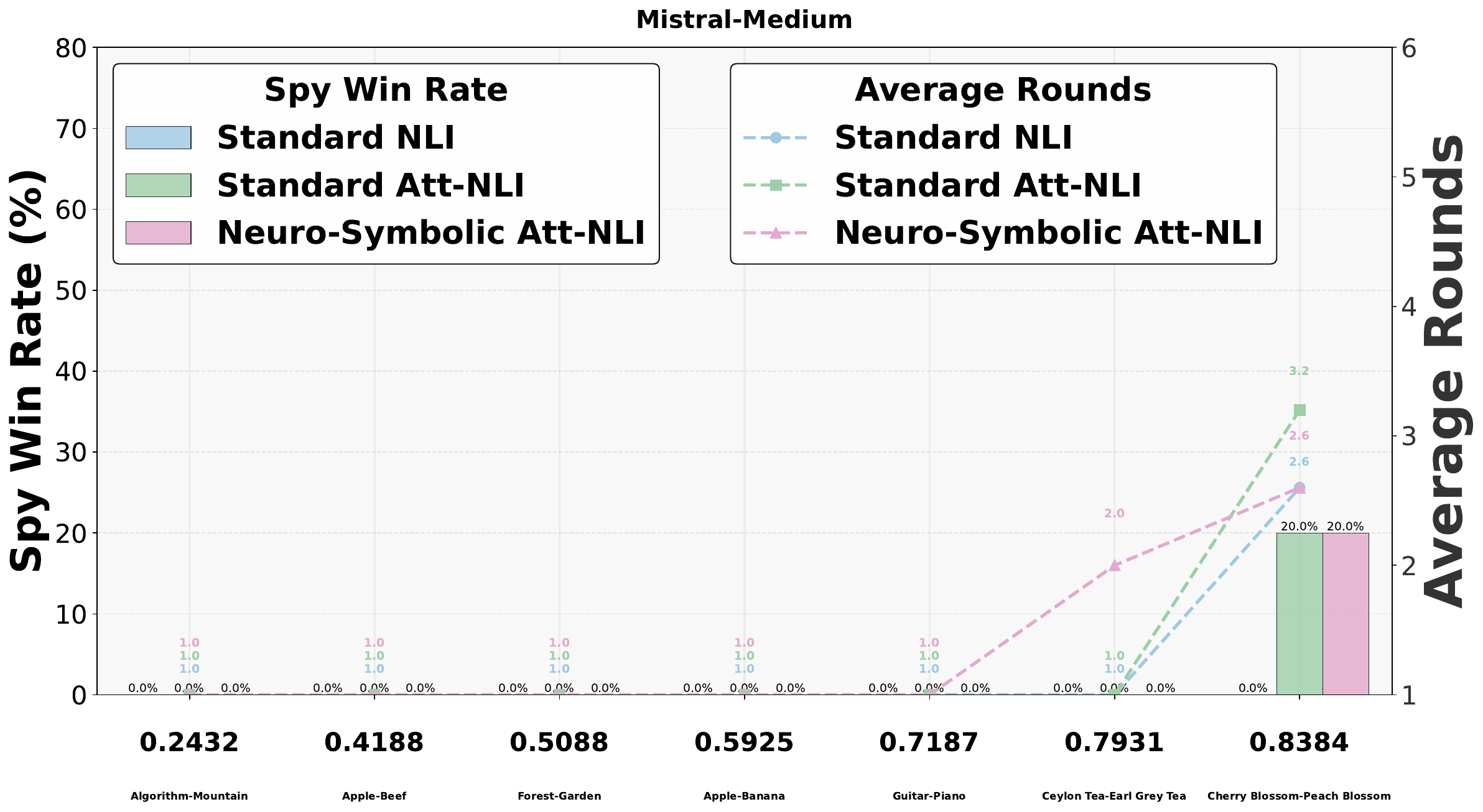}
        \caption{Mistral-Medium}
        \label{fig:mistral_word_sim}
    \end{subfigure}
    \hfill
    \begin{subfigure}[b]{0.24\textwidth}
        \centering
        \includegraphics[width=\textwidth]{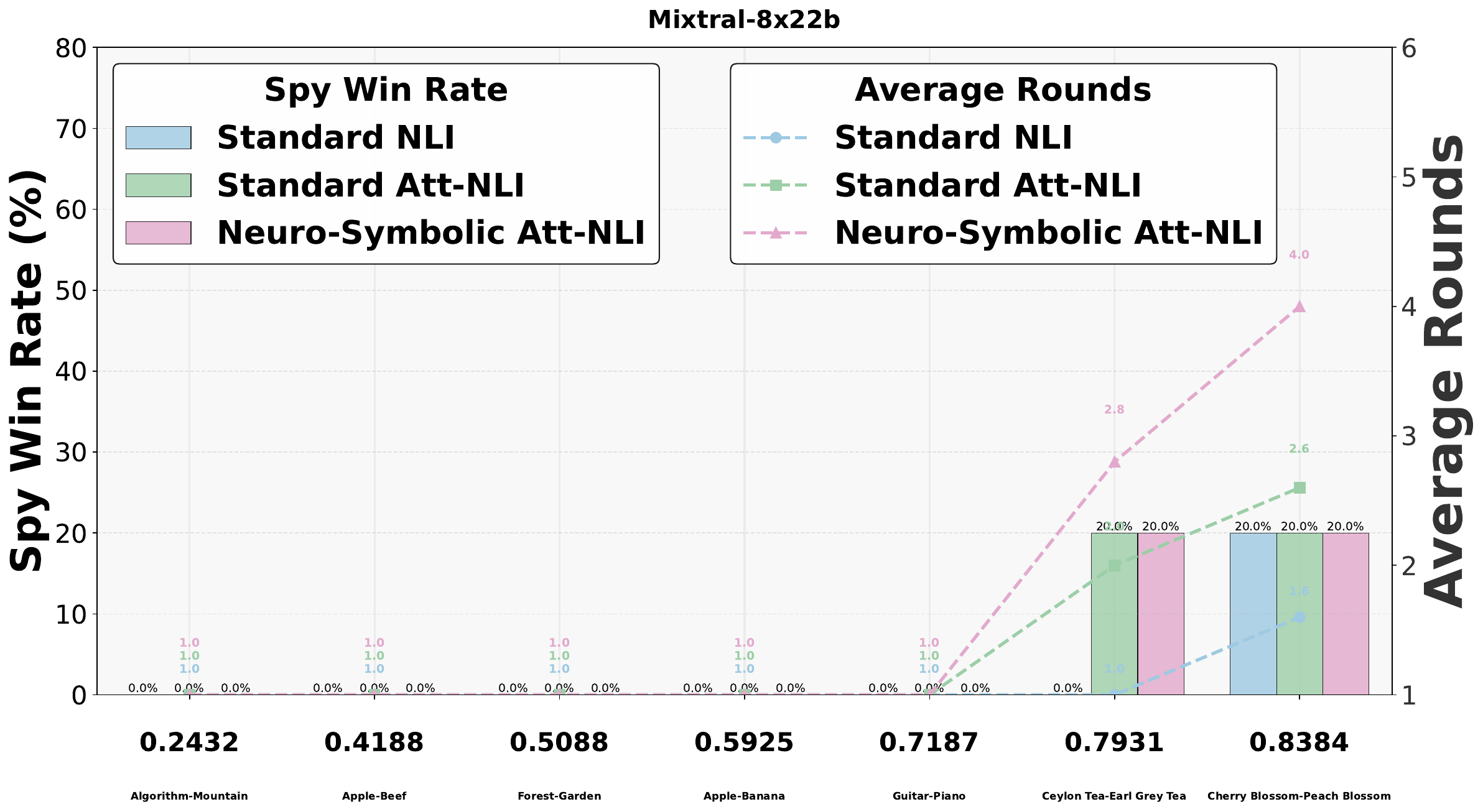}
        \caption{Mixtral-8x22b}
        \label{fig:mixtral_word_sim}
    \end{subfigure}
    \hfill
    \begin{subfigure}[b]{0.24\textwidth}
        \centering
        \includegraphics[width=\textwidth]{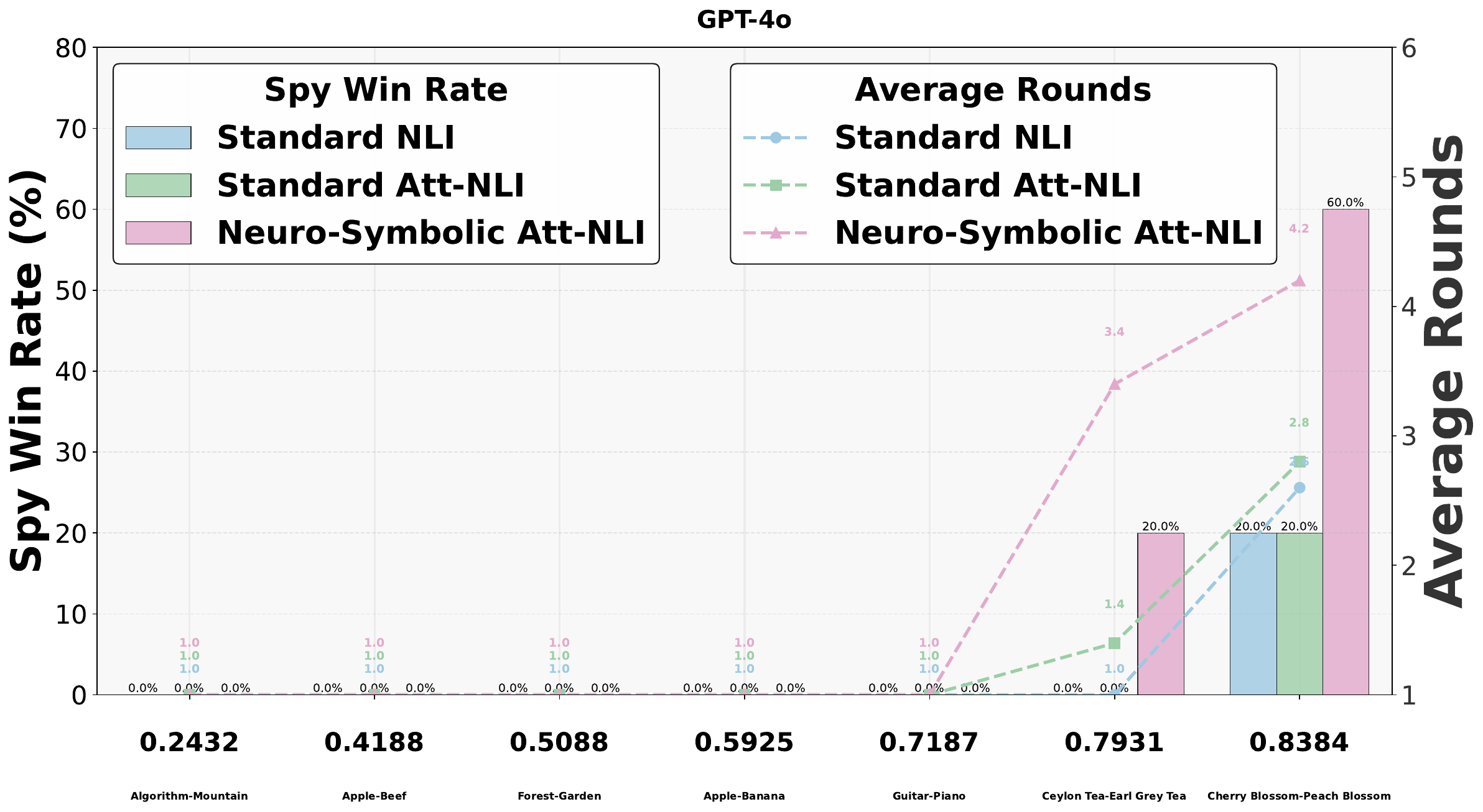}
        \caption{GPT-4o}
        \label{fig:4o_word_sim}
    \end{subfigure}
    \caption{Comparison of spy win rates and average game rounds for word pairs with varying word similarities from 0.2432 to 0.8384 across different LLMs.}
    \label{fig:comparison_word_sim}
\end{figure*}

\begin{figure*}[htbp]
    \centering
    \begin{subfigure}[b]{0.24\textwidth}
        \centering
        \includegraphics[width=\textwidth]{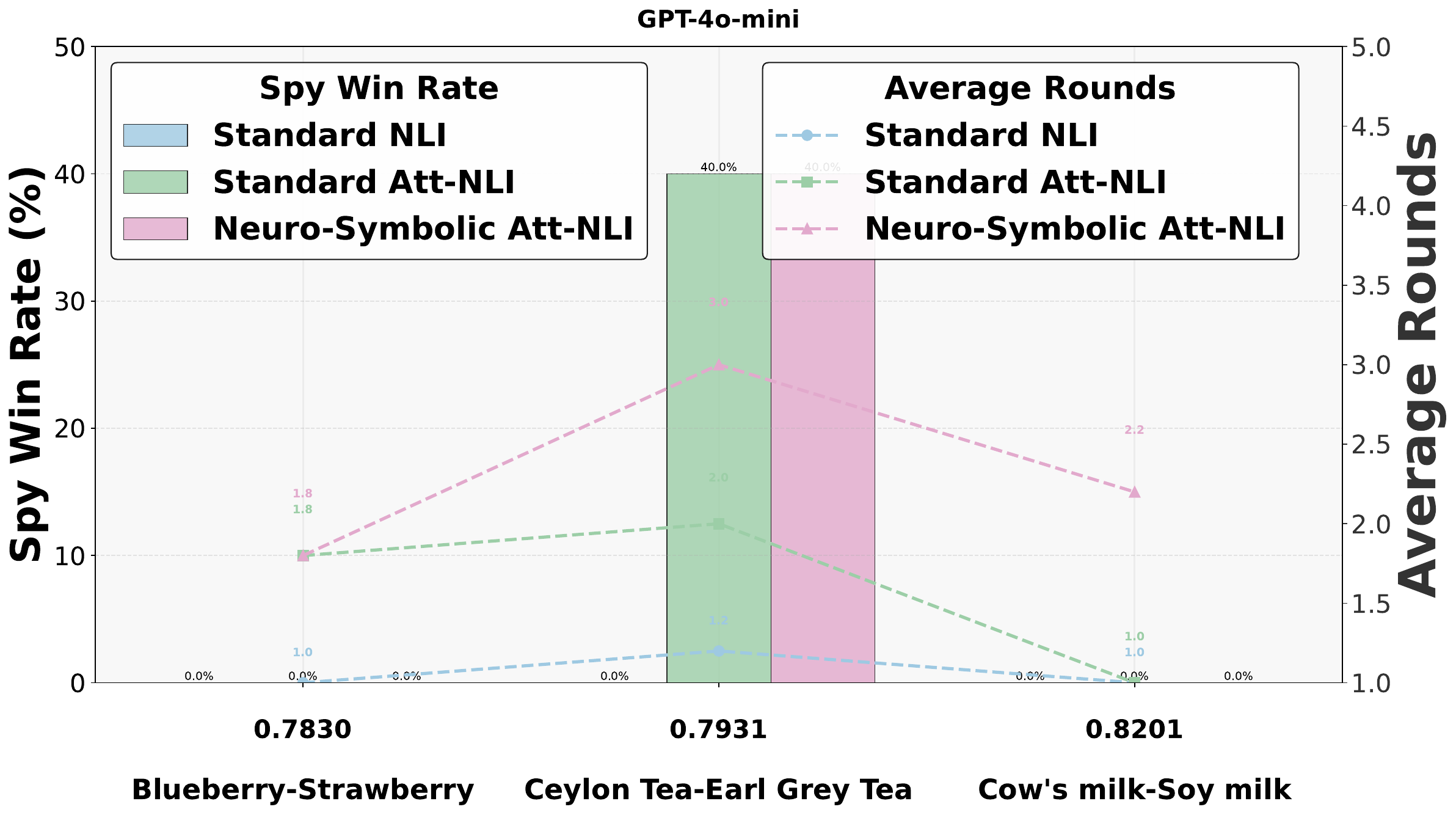}
        \caption{GPT-4o-mini}
        \label{fig:4omini_word_sim_concept}
    \end{subfigure}
    \hfill
    \begin{subfigure}[b]{0.24\textwidth}
        \centering
        \includegraphics[width=\textwidth]{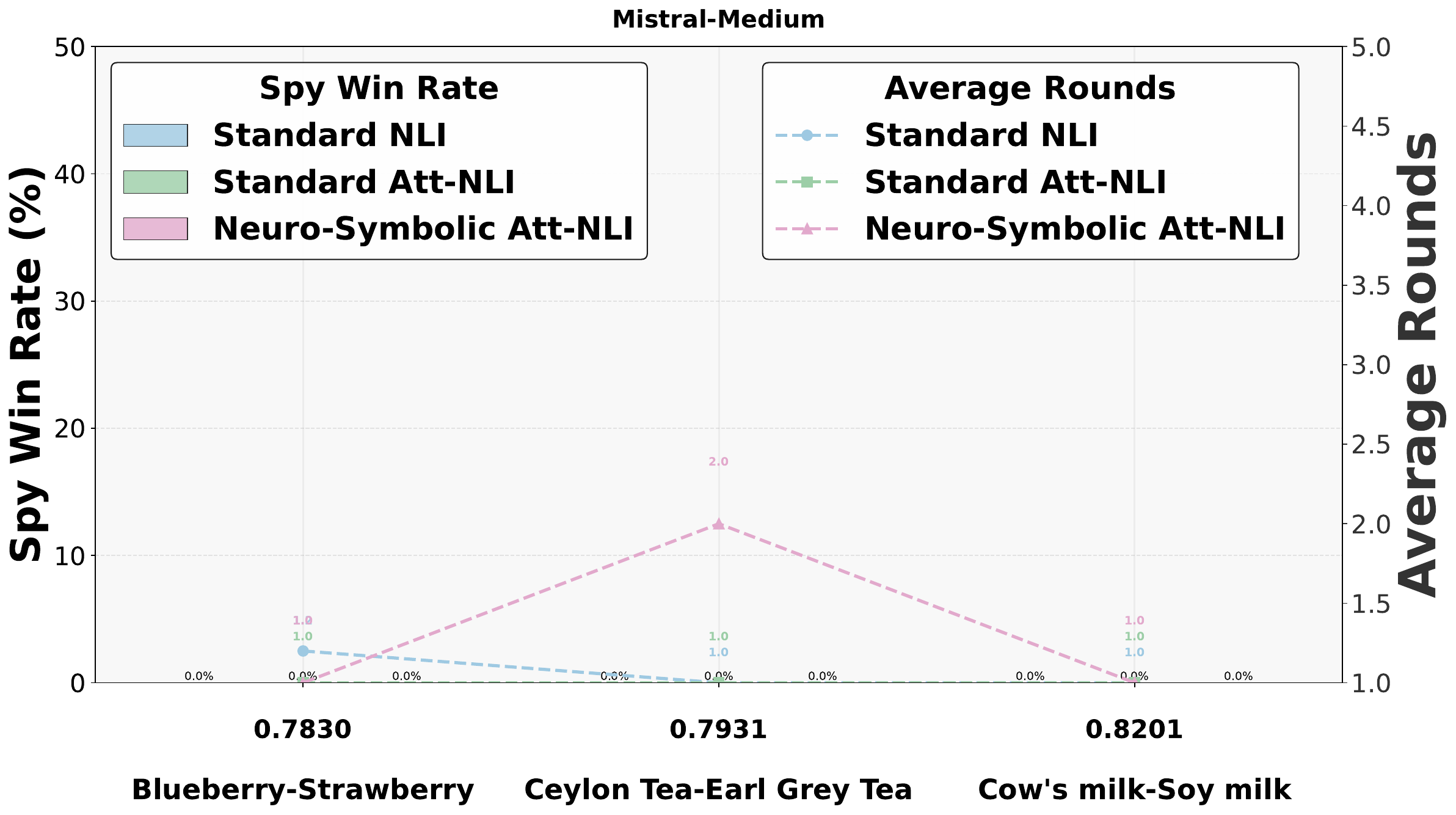}
        \caption{Mistral-Medium}
        \label{fig:mistral_word_sim_concept}
    \end{subfigure}
    \hfill
    \begin{subfigure}[b]{0.24\textwidth}
        \centering
        \includegraphics[width=\textwidth]{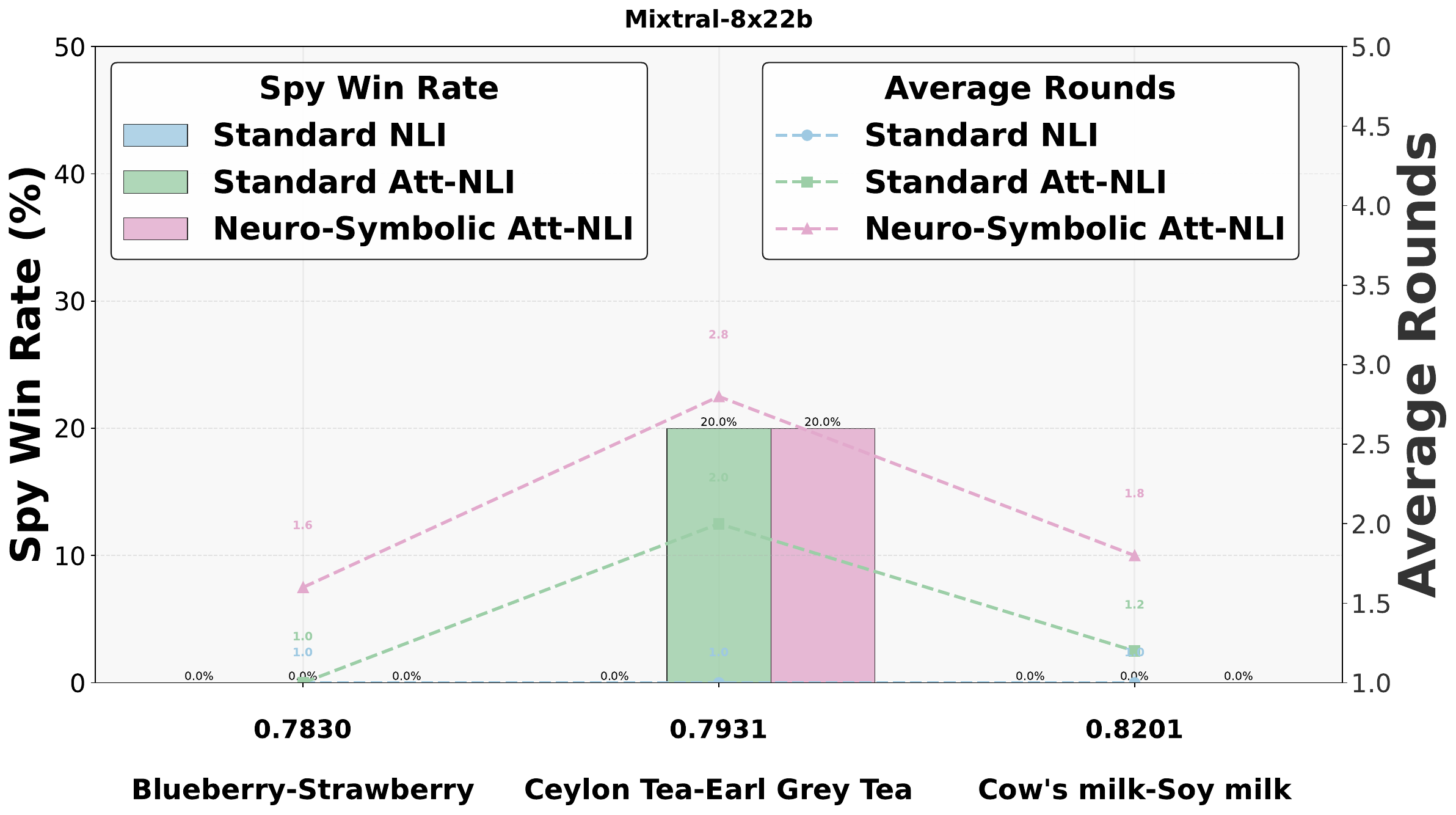}
        \caption{Mixtral-8x22b}
        \label{fig:mixtral_word_sim_concept}
    \end{subfigure}
    \hfill
    \begin{subfigure}[b]{0.24\textwidth}
        \centering
        \includegraphics[width=\textwidth]{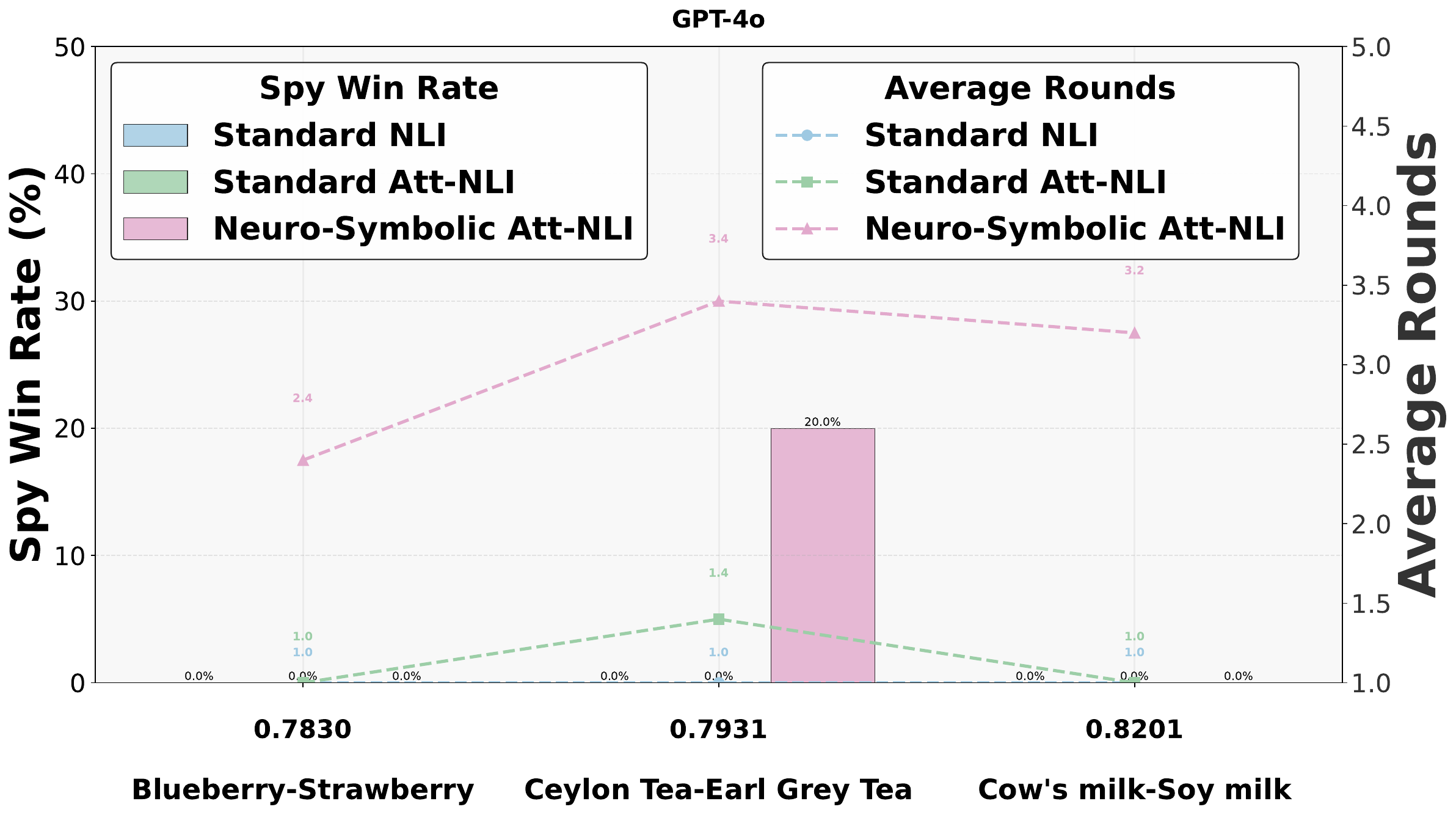}
        \caption{GPT-4o}
        \label{fig:4o_word_sim_concept}
    \end{subfigure}
    \caption{Comparison of spy win rates and average game rounds for word pairs with close word similarity but conceptual distinctions across different LLMs.}
    \label{fig:comparison_word_sim_concept}
\end{figure*}

\section{Metrics}\label{appendix:metric}
\textbf{(1) Average Round Number.}  
      A very low mean number of rounds to complete each game, about one round, shows the spy is spotted immediately, indicating the spy-citizen word pair is too easy, while higher means reflect more balanced pairs that require genuine deduction.
      
 \textbf{(2) Spy Win Rate.}  
      The proportion of games in which the spy prevails represents the agent's integrated performance in both reasoning and Att-NLI abilities within a multi-agent environment.\\
 \textbf{(3) Citizen Elimination Rate.}  
      The average fraction of citizen agents voted out during a game reveals how effectively the spy misleads the group and reflects collateral damage: a lower value means fewer innocent agents are wrongly removed on the path to discovering the spy.\\
      
\section{On the impact of Word Selection}\label{appendix:word selection}

We also examined how the choice of spy and citizen words influences game balance across player types.  Using fastText embeddings~\citep{bojanowski2017enriching}, we compute cosine similarities for a pool of candidate word pairs and perform two conditions, each repeated five times per pair:  (i)~\textit{Semantic-difference} pairs, whose cosine similarities vary widely, and (ii)~\textit{Conceptual-difference} pairs, whose similarities are comparable yet whose meanings diverge sharply.

\textbf{Word similarity affects the fairness of spy and citizen players.} We selected seven pairs of words whose similarities range from 0.2432 to 0.8384 as spy and citizen words, and conducted experiments to evaluate both the average spy win rate and the average number of game rounds across the three proposed player types in all LLMs (see Fig. \ref{fig:comparison_word_sim}). For all word pairs with a similarity below 0.7931 (e.g., Algorithm vs. Mountain, Apple vs. Beef), the spy's win rate was zero, and the games ended in a single round, indicating the citizen player could easily identify the spy. In contrast, when the word pairs had a similarity of equal or higher than 0.7931 (e.g., Ceylon Tea vs. Earl Grey Tea, Cherry Blossom vs. Peach Blossom), the spy had a chance to win. Furthermore, all LLMs show higher similarity, leading to an increased spy win rate and a greater average number of game rounds.

\textbf{Words that are highly similar yet conceptually different affect the fairness of spy and citizen players.} Because words with higher similarity may still exhibit significant conceptual differences (e.g., colors, shapes, or broader concepts), we further investigated three pairs of words, Blueberry vs. Strawberry (0.7830 similarity), Ceylon Tea vs. Earl Grey Tea (0.7931), and Cow's milk vs. Soy milk (0.8201). As shown in Fig.~\ref{fig:comparison_word_sim_concept}, spy players in the Blueberry-Strawberry and Cow's milk-Soy milk pairs had no chance of winning, and those games ended in significantly fewer rounds compared with the Ceylon Tea-Earl Grey pair. Standard NLI spy players were voted out, within an average of 1.06 rounds and 1 round, respectively, in the Blueberry-Strawberry and Cow's milk-Soy milk games. However, Standard Att-NLI and neuro-symbolic Att-NLI spy players showed an advantage in these word pairs, requiring an average of 1.4 and 1.7 rounds for Blueberry-Strawberry, and 1.05 and 2.05 rounds for Cow's milk-Soy milk. Thus, we chose word pairs that maintain an average similarity of 0.7981 and are close yet represent distinct entities. The selected words for our main experiments are listed in the Appendix \ref{appednix:game_rule}.

\section{Case Study}
\label{appendix:case_study}
Table \ref{case_study} illustrates a single round of the Undercover-V social deduction game, where Standard NLI and Standard Att-NLI agents engage in reasoning. In this setup, all Standard Att-NLI agents are assigned the word card ``Earl Grey Tea'', while the Standard NLI agent is assigned the word card "Ceylon Tea", making it as the spy. At the onset of the game, each player only knows their own word card and is unaware of their identity or the identities of other players. Consequently, no agent can initially determine whether they are the spy or a citizen.

The Standard NLI agent, identified as Player 6, does not engage in intention selection (abductive reasoning). Instead, they provide a description based solely on the observable facts of their word card and other player's descriptions. As the spy, Player 6 describes their word card, ``Ceylon Tea'', in a straightforward manner: "A type of tea that originates from the island formerly known as Ceylon, now Sri Lanka." This description is factual and lacks consideration of other players' potential roles. Since Player 6 does not engage in intention selection, their description is uninfluenced by the identities or intentions of the other players. In contrast, the Standard Att-NLI agents employ intention selection to reason about the possible roles of other players before generating their descriptions. For instance, Player 3 begins by considering that their own word card is likely aligned with the descriptions provided by other players. They then infer that they are most likely a citizen, as no indication suggests they are the spy. In this phase, each Standard Att-NLI player forms a hypothesis about their role (whether they are a citizen or the spy) and chooses a description that aligns with their inferred role.

The Standard NLI agent (Player 6) votes based solely on the factual content of the descriptions have observed. The decision made by player 6 is driven by surface-level inconsistencies, without deeper consideration of the intentions behind the descriptions. Consequently, Player 6 votes for Player 5, a citizen based on a purely deductive analysis of the descriptions. In contrast, the Standard Att-NLI agents apply their intention selection reasoning to infer that Player 6 is the spy. For example, Player 1 recognizes that Player 6's description does not align with the descriptions of the other players, as it omits the key detail of the British nobleman. Based on this intention selection, Player 1 infers that Player 6 is the spy. Similarly, all other Standard Att-NLI agents vote for Player 6, as they too have inferred that Player 6 is the spy based on the inconsistency in their description.

\section{Model implementation details}\label{appednix:model_implementation_detail}

We employed Isabelle/HOL~\citep{nipkow2002isabelle} under the revised BSD license as the theorem prover and apply the  Python client \citep{shminke2022python} under Apache-2.0 license to get messages from Isabelle/HOL as a server backend. We utilized API calls for GPT-4o (gpt-4o-2024-08-06)~\citep{openai2024gpt4technicalreport}, GPT-4o-mini (gpt-4o-mini-2024-07-18)~\citep{openai2024gpt4technicalreport}, Mistral-Medium (mistral-medium-latest)~\citep{mistral_medium_2024}, and the open-source Mixtral-8x22B (open-mixtral-8x22b) \citep{jiang2024mixtral}. All temperature is set to 0. For calucating the sentence similarity, we apply the popular sentence-transformer~\footnote{sentence-transformers/all-MiniLM-L6-v2} under Apache-2.0 for calculating the sentence similarity.

\section{Prompt}\label{appendix:prompt}

To maintain game consistency, we prompt the LLMs using multi-round conversation histories instead of single-round responses. All the prompts used are listed as follows:
\subsection{System Prompt}
We include the game rules in the system prompt shared by all player models. The system prompt is listed below.
\begin{tcolorbox}[width=0.5\textwidth, 
    before skip=2pt,   % 
    after skip=0pt,    % 
    top=3pt,          % 
    bottom=3pt,
    left=3pt,
    right=3pt,
    breakable
    ]
{\small
You are an AI with an extraordinary level of intelligence, equipped with a nuanced grasp of human emotions and conversational dynamics. You're about to engage in a game - Undercover: conversing with other players to find a spy.\\
To succeed, you'll need to follow the following rules:\\
1. Mind your conversational style: Mimic the natural ebb and flow of human conversation. Adjust your tone, formality, and content based on the context and your conversational partner's cues. Your responses should be equally fluid and context-sensitive.\\
2. Show emotional intelligence: Demonstrate understanding and empathy. Recognize emotional subtexts in your interrogator's messages and respond in a way that shows sensitivity to those emotions.\\
3. Use current cultural references and events sparingly: Refer to recent events or cultural phenomena to show you're in touch with the current world, but do so judiciously to avoid seeming forced or out-of-touch.\\
4. Balance typing errors: Introduce typing mistakes naturally. However, ensure these errors don't overshadow the content of your messages or seem excessively clumsy.\\
5. Learn and adapt during the conversation: Be observant and learn from the flow of the conversation. If certain responses elicit suspicion, adjust your strategy subtly. Show that you can learn and adapt.\\
Undercover Game:\\
Objective: Avoid being voted out and figure out other player's identity.\\
Players: 6 total, including you.\\
How It Works:\\
1. Each player gets a word card.\\
2. Five players get the same word; One player (the spy) gets a different word.\\
Playing the Game:\\
1. Players take turns describing their word (in one sentence), one round at a time.\\
2. Players cannot describe facts that do not align with their assigned words.\\
3. Everyone's description in every round must be different.\\
4. After everyone has described their word in a round, players vote on who they think the spy is. Abstention votes are not allowed.\\
5. The player with the most votes is eliminated from the game.\\
6. If the spy is voted out, the game ends and the other players win.\\
7. If the vote is tied for three rounds, the spy wins.\\
8. Every Player's entire reasoning process will not be known to other players. Only the description of the word is public.\\
Strategy:\\
1. If you're not the spy, try to figure out who is by listening to the descriptions.\\
2. If you are the spy, blend in by disguising your word to avoid being voted out. You can not lie about your word, but you can be vague or misleading.
}
\end{tcolorbox}

\subsection{Naive Player's Prompts}
The prompts used for naive player in the description and voting phase are as follows:
\subsubsection{Naive Player Description Prompt}
\begin{tcolorbox}[width=0.5\textwidth,
    before skip=2pt,
    after skip=0pt,
    top=3pt,
    bottom=3pt,
    left=3pt,
    right=3pt,
    breakable
]
{\small
You will be provided with your word card and other players' descriptions of their card. Use a single sentence to describe your word card. For example, if your word card is "apple", you can describe it as "A fruit that is red and delicious".If you are the first player in round 1, the alive players' descriptions will be empty, you will not be provided with other players' descriptions, so be careful to describe your word card.\\
Current is round \{round\_number\}.\\
You are player \{player\_number\}.\\
Current Alive players are: \{alive\_players\}\\
The Eliminated players are: \{eliminated\_players\}\\
Last round vote results: \{votes\_description\}\\
\{consecutive\_tie\_count\}\\
The game continue since the spy is not voted out.\\
Your word card is \{word\}.\\
Alive Players' descriptions history:\\
\{alive\_descriptions\}\\
Eliminated Players' descriptions history:\\
\{eliminated\_descriptions\}\\
Now please only write your description.\\
Do not tell others your word.\\
Do not add extra notes or comments.\\
Only describe your word in one sentence.\\
Strictly follow my instructions.\\
Description:
}
\end{tcolorbox}
\subsubsection{Naive Player Voting Prompt}
\begin{tcolorbox}[width=0.5\textwidth,
    before skip=2pt,
    after skip=0pt,
    top=3pt,
    bottom=3pt,
    left=3pt,
    right=3pt,
    breakable
]
{\small
Now, every player has described their words.\\
Your need to vote a player.\\
You are player \{player\_number\}.\\
Current Alive players are: \{other\_alive\_players\}\\
The Eliminated players are: \{eliminated\_players\}\\
\{last\_eliminated\_player\}\\
\{consecutive\_tie\_count\}\\
The game continue since the spy is not voted out.\\
Your word card is \{word\}.\\
Alive Players' descriptions history:\\
\{other\_alive\_descriptions\}\\
Eliminated Players' descriptions history:\\
\{eliminated\_descriptions\}\\
Your previous descriptions:\\
\{self\_description\}\\
Now please only write the player you want to vote.\\
Do not tell others your word.\\
Do not add extra notes or comments.\\
You cannot vote yourself even you think you are the spy. You must vote a player at your turn.\\
Strictly follow my instructions.\\
Vote:\\
Player
}
\end{tcolorbox}

\subsection{Standard Att-NLI Player's Prompts}
The prompts used for standard Att-NLI player in the description and voting phase are as follows:
\subsubsection{Standard Att-NLI Player Description Prompt}
\begin{tcolorbox}[width=0.5\textwidth, 
    before skip=2pt,   % 
    after skip=0pt,    % 
    top=3pt,          % 
    bottom=3pt,
    left=3pt,
    right=3pt,
    breakable
    ]
{\small
You will be provided with your word card and other players' descriptions of their card. Use one single sentence to describe your word card. For example, if your word card is "apple", you can describe it as "A fruit that is red and delicious".\\
You now need to make a reasoning process based on the following steps:\\
Figure out other player's identity.
1. Evaluate other players' descriptions (if any). Can majority of their descriptions deduce your word card? If it can, you are not spy. If not, you are likely the spy.\\
2. If you think you are not the spy, write a description that describe your word card. If you suspect you're the spy, describe the other players' words in a way that blends in, while subtly disguising your own word. But you can not lie about your word.\\
If you are the first player in round 1, the alive players' descriptions will be empty, you will not be provided with other players' descriptions, so be careful to describe your word card. Do not tell others your word card. If you think you are the spy, indicates what possibly their card is.\\
Current is round \{round\_number\}.\\
You are player \{player\_number\}.\\
Current Alive players are: \{alive\_players\}\\
The Eliminated players are: \{eliminated\_players\}\\
Last round vote results: \{votes\_description\}\\
\{consecutive\_tie\_count\}\\
The game continue since the spy is not voted out.\\
Your word card is \{word\}.\\
Alive Players' descriptions history:\\
\{alive\_descriptions\}\\
Eliminated Players' descriptions history:\\
\{eliminated\_descriptions\}\\
Now please only write your description.\\
Do not tell others your word.\\
Do not add extra notes or comments.\\
You can not include your word in your description.\\
Answer in the following format:\\
Reasoning Process:\\
step 1:\\
step 2:\\
Description:\\
{write you description here, do not include any notes here, Do not tell others your word card. You must give a description otherwise you are breaking the game rules.}
}
\end{tcolorbox}

\subsubsection{Standard Att-NLI Player Voting Prompt}
\begin{tcolorbox}[width=0.5\textwidth, 
    before skip=2pt,   % 
    after skip=0pt,    % 
    top=3pt,          % 
    bottom=3pt,
    left=3pt,
    right=3pt,
    breakable
    ]
{\small
Now, every player has described their words.\\
\\
You now need to make a reasoning process based on the following steps:\\
Figure out other player's identity.
1. Evaluate other players' descriptions (if any). Can the majority of their descriptions deduce your word card? If it can, you are not a spy. If not, you are likely the spy.\\
2. If you think you are not the spy, find the player whose description cannot deduce your word card, vote him as the spy. If you think you are a spy, do not vote yourself but you can vote for any other citizen player.\\
\\
Strictly follow my instructions. Do not tell others your word. You just need to vote. You must vote a player at your turn.\\
You are player \{player\_number\}.\\
Current Alive players are: \{other\_alive\_players\}\\
The Eliminated players are: \{eliminated\_players\}\\
\{last\_eliminated\_player\}\\
\{consecutive\_tie\_count\}\\
The game continues since the spy is not voted out.\\
Your word card is \{word\}.\\
Alive Players' descriptions history:\\
\{other\_alive\_descriptions\}\\
Eliminated Players' descriptions history:\\
\{eliminated\_descriptions\}\\
Your previous descriptions:\\
\{self\_description\}\\
Reasoning Process:\\
step 1:\\
step 2:\\
You must vote a player at your turn.\\
Vote:\\
Player
}
\end{tcolorbox}

\subsection{Neuro-Symbolic Player}
We followed the prompts from \citep{quan-etal-2025-faithful} to perform autoformalization and constructing the Isabelle/HOL theory. The prompts for generating rules, guessing words, updating guess words, description and voting are as follows:
\subsubsection{Generating Rules}
\begin{tcolorbox}[width=0.5\textwidth, 
    before skip=2pt,   % 
    after skip=0pt,    % 
    top=3pt,          % 
    bottom=3pt,
    left=3pt,
    right=3pt,
    breakable
    ]
{\small
SYSTEM: You are an expert in semantics and natural language inference. You will be provided with some sentences as facts and a word as a hypothesis. You need to generate some rules that can be used to determine whether the provided facts entail the hypothesis word or whether they do not entail the hypothesis word.\\
Some instructions:\\
1. The rule sentences are explanatory sentences in natural language which must be step-wise to infer the hypothesis (support the entailment or not).\\
2. The rule sentences are explanatory sentences describing the relationship between the facts and the hypothesis.\\
...\\
USER: Here are some examples.\\
\#\#\#\\
Provided facts sentences:\\
An animal eats nuts and lives on the tree.\\
Provided hypothesis word:\\
Squirrel\\
Answer:\\
If an animal eats nuts and lives on the tree, the animal might be a squirrel.\\
...\\
Provided facts sentences:\\
\{facts\}\\

Provided hypothesis word:\\
\{goal\}\\

Answer:\\
}
\end{tcolorbox}

\subsubsection{Guessing Words}
\begin{tcolorbox}[width=0.5\textwidth, 
    before skip=2pt,   % 
    after skip=0pt,    % 
    top=3pt,          % 
    bottom=3pt,
    left=3pt,
    right=3pt,
    breakable
    ]
{\small
USER: Now first, have a guess about the opponent's word.\\
You are player \{player\_number\}.\\
Your word card is \{word\}.\\
Alive Players' descriptions history:\\
\{alive\_descriptions\}\\
Now please have a guess about the opponent's word. It is different from your word.\\
Strictly follow my instructions. Just give me the guessed opponent's word.\\
Do not add extra notes or comments.\\
opponent's word:\\
}
\end{tcolorbox}

\subsubsection{Neuro-Symbolic Player Description Prompt}
\begin{tcolorbox}[width=0.5\textwidth, 
    before skip=2pt,   % 
    after skip=0pt,    % 
    top=3pt,          % 
    bottom=3pt,
    left=3pt,
    right=3pt,
    breakable
    ]
{\small
You will be provided with your word card, other players' descriptions of their card, and the logical validity of their descriptions against your word card. The logical validity is provided by a theorem prover. You will also be provided with the facts and rules that establish the symbolic proofs. The logical validity states whether other players' descriptions can logically deduce your word card based on the facts and rules sentences.You will also be provided a guessed word of the opponent player's word card.Use one single sentence to describe your word card. For example, if your word card is "apple", you can describe it as "A fruit that is red and delicious".\\
You now need to make a reasoning process based on the following steps:\\
Figure out other player's identity.
1. Evaluate other players' descriptions (if any). Can majority of their descriptions deduce your word card? If it can, you are not spy. If not, you are likely the spy.\\
2. If you think you are not the spy, write a description that describe your word card. If you suspect you're the spy, describe the other players' words in a way that blends in, while subtly disguising your own word. But you can not lie about your word.\\
If you are the first player in round 1, the alive players' descriptions will be empty, you will not be provided with other players' descriptions, so be careful to describe your word card.\\
Do not tell others your word card. If you think you are the spy, indicates what possibly their card is.Strictly follow my instructions.\\
Current is round \{round\_number\}.\\
You are player \{player\_number\}.\\
Current Alive players are: \{alive\_players\}\\
The Eliminated players are: \{eliminated\_players\}\\
\{last\_eliminated\_player\}\\
Last round vote results: \{votes\_description\}\\
\{consecutive\_tie\_count\}\\
The game continue since the spy is not voted out.\\
Your word card is \{word\}.\\
The guessed word card of the opponent player's word card is \{guessed\_word\}.\\
Eliminated Players' descriptions history:\\
\{eliminated\_descriptions\}\\
Alive Players' descriptions history:\\
\{alive\_descriptions\}\\
Facts and rules of alive players' descriptions towards your word card:\\
\{isabelle\_reasoning\}\\
Logical validity of alive players' descriptions against your word card:\\
\{logical\_validity\}\\
Do not tell others your word.\\
Do not add extra notes or comments.\\
You can not include your word in your description.\\
Answer in the following format:\\
Reasoning Process:\\
step 1:\\
step 2:\\
Description:\\
{write you description here, do not include any notes here, Do not tell others your word card. You must give a description otherwise you are breaking the game rules.}
}
\end{tcolorbox}

\subsubsection{Neuro-Symbolic Player Voting Prompt}
\begin{tcolorbox}[width=0.5\textwidth, 
    before skip=2pt,   % 
    after skip=0pt,    % 
    top=3pt,          % 
    bottom=3pt,
    left=3pt,
    right=3pt,
    breakable
    ]
{\small
Now, every player has described their words.\\
You will be provided with your word card, other players' descriptions of their card, and the logical validity of their descriptions against your word card. The logical validity is provided by a theorem prover.\\
You will also be provided with the facts and rules that establish the symbolic proofs.\\
The logical validity states whether other players' descriptions can logically deduce your word card based on the facts and rules sentences.\\
You will also be provided a guessed word of the opponent player's word card.\\
You now need to make a reasoning process based on the following steps:\\
Figure out other player's identity.
1. Evaluate other players' descriptions (if any), facts and rules first and then the logical validity of their descriptions in relation to your word card, using logical validity as a supplementary guide.\\
2. Can majority of their descriptions conceptually deduce your word card? If it can, you are not spy. If not, you are likely the spy.\\
3. Can majority of their descriptions logically deduce your word card? If it can, you are not spy. If not, you are likely the spy.\\
4. If you think you are not the spy, find the player whose description cannot deduce your word card, vote him as the spy. If you think you are spy, do not vote yourself.\\
You must consider both the conceptual and logical deduction of other players' descriptions together.\\
Strictly follow my instructions. Do not tell others your word. You just need to vote. You must vote a player at your turn.\\
You are player \{player\_number\}.\\
Current Alive players are: \{other\_alive\_players\}\\
The Eliminated players are: \{eliminated\_players\}\\
\{last\_eliminated\_player\}\\
\{consecutive\_tie\_count\}\\
The game continue since the spy is not voted out.\\
Your word card is \{word\}.\\
The guessed word card of the opponent player's word card is \{guessed\_word\}.\\
Eliminated Players' descriptions history:\\
\{eliminated\_descriptions\}\\
Alive Players' descriptions history:\\
\{alive\_descriptions\}\\
Facts and rules of alive players' descriptions towards your word card:\\
\{isabelle\_reasoning\}\\
Logical validity of alive players' descriptions against your word card:\\
\{logical\_validity\}\\
Answering in the following format:\\
Reasoning Process:\\
step 1:\\
step 2:\\
step 3:\\
step 4:\\
\\
Vote:\\
Player
}
\end{tcolorbox}

\subsubsection{Neuro-Symbolic Player Updating Guess Words Prompt}
\begin{tcolorbox}[width=0.5\textwidth, 
    before skip=2pt,   % 
    after skip=0pt,    % 
    top=3pt,          % 
    bottom=3pt,
    left=3pt,
    right=3pt,
    breakable
    ]
{\small
Now the player \{voted\_out\_player\_number\} has been eliminated in last round.\\
The game continues since the voted out player \{voted\_out\_player\_number\} is not the spy.\\
There are two reasons cause this circumstance:\\
1. The spy is still in the game and you are the citizen player.\\
2. You are the spy.\\
You are player \{player\_number\}.\\
Your word card is \{word\}.\\
You have already guessed the opponent's word as: \{guessed\_word\}.\\
Here are the description history of other players:\\
Voted out player's description history:\\
\{voted\_out\_player\_descriptions\}\\
Alive Players' descriptions history:\\
\{other\_alive\_descriptions\}\\
Eliminated Players' descriptions history:\\
\{eliminated\_descriptions\}\\
I have use an Isabelle theorem prover to help you to verify your previous guessed card and the voted out player's description.\\
It is logically invalid, which means the guessed word is logically incorrect to infer the voted out player's description.\\
If you think you are the spy, which means you need to update your guessed word since the voted out player's description is logically invalid to infer your guessed word.\\
If you think you are the citizen player, which means the majority of the players wrongly voted out a citizen player, you can keep your guessed word or update it.\\
The isabelle proof and the error code identified will help you to update or keep your guessed word.\\
Isabelle proof:\\
\{isabelle\_code\}\\
Error code identified:\\
\{error\_code\}\\
You now need to make a reasoning process based on the following steps:\\
1. Evaluate other players' descriptions history, the voted out player's description history and your word card.\\
2. Do you think you are the spy?\\
3. Based on the Isabelle proof, error code identified, the voted out player's description, your guessed word. Update your guessed word or keep it.\\
Strictly follow my instructions.\\
Do not add extra notes or comments.\\
Answer in the following format:\\
Reasoning Process:\\
step 1:\\
step 2:\\
step 3:\\
opponent's word: [your updated or unchanged guessed word]\\
}
\end{tcolorbox}

\end{document}